# Dempster's Rule for Evidence Ordered in a Complete Directed Acyclic Graph

**Ulla Bergsten and Johan Schubert**
*Division of Applied Mathematics and Scientific Data Processing,*
*Department of Weapon Systems, Effects and Protection,*
*National Defence Research Establishment, Sundbyberg, Sweden.*

## ABSTRACT

*For the case of evidence ordered in a complete directed acyclic graph this paper presents a new algorithm with lower computational complexity for Dempster's rule than that of step-by-step application of Dempster's rule. In this problem, every original pair of evidences, has a corresponding evidence against the simultaneous belief in both propositions. In this case, it is uncertain whether the propositions of any two evidences are in logical conflict. The original evidences are associated with the vertices and the additional evidences are associated with the edges. The original evidences are ordered, i.e., for every pair of evidences it is determinable which of the two evidences is the earlier one. We are interested in finding the most probable completely specified path through the graph, where transitions are possible only from lower- to higher-ranked vertices. The path is here a representation for a sequence of states, for instance a sequence of snapshots of a physical object's track. A completely specified path means that the path includes no other vertices than those stated in the path representation, as opposed to an incompletely specified path that may also include other vertices than those stated. In a hierarchical network of all subsets of the frame, i.e., of all incompletely specified paths, the original and additional evidences support subsets that are not disjoint, thus it is not possible to prune the network to a tree. Instead of propagating belief, the new algorithm reasons about the logical conditions of a completely specified path through the graph. The new algorithm is $O(|\Theta| \log |\Theta|)$, compared to $O(|\Theta|^{\log |\Theta|})$ of the classic brute force algorithm. After a detailed presentation of the reasoning behind the new algorithm we*

---

*Address correspondence to Johan Schubert, Division of Applied Mathematics and Scientific Data Processing, Department of Weapon Systems, Effects and Protection, National Defence Research Establishment, S-172 90 Sundbyberg, Sweden.*

*conclude that it is feasible to reason without approximation about completely specified paths through a complete directed acyclic graph.*

## 1. INTRODUCTION

The development of knowledge-based systems has evoked increasing attention to the subject of approximate reasoning. The available information in a system is often uncertain, incomplete, and even partly incorrect–demanding methods able to handle this kind of information. The Dempster-Shafer theory, which provides an attractive representation of uncertainty and an intuitive combination of uncertain information, is one such method (Dempster [1], Shafer [2, 3, 4]). However, one problem with the Dempster-Shafer theory is its computational complexity. In many cases even a moderate amount of data leads to huge computational complexity making it necessary either to aggregate focal elements, i.e., use summarization (Lowrance et al. [5]), or to derive approximate or special case algorithms.

In this paper we present an algorithm for the special case of evidences ordered in a complete directed acyclic graph. In this case, it is uncertain whether the propositions of any two evidences are in logical conflict. Here, we can model the uncertainty by an additional evidence against the simultaneous belief in both propositions and treat the two original propositions as non-conflicting. This will give rise to a complete directed acyclic graph with the original evidences on the vertices and the additional ones on the edges. As an example, we may think of the vertices as positions in time and space and the edges as transitions between these positions. Transitions are only possible from a vertex with a lower index to one with a higher. We are interested in finding the most probable path of an object. The evidence at a vertex may then be an evidence that the object has been at that position and the evidence at an edge an evidence against the possibility of a transition between the two positions. The classic algorithm calculates the support and plausibility for a given path, i.e., a sequence of vertices, through the graph by first combining all evidences step-by-step with Dempster's rule and then summing up all contributions for the path. The new algorithm reasons instead about the logical conditions of a completely specified path through the complete directed acyclic graph, gaining significantly in time and space complexity.

In this paper, we give a brief summary of Dempster-Shafer theory (Section 2), discuss the type of problem domains that satisfy our restrictions and then describe the representation of Dempster-Shafer theory in this case (Section 3). In Section 4 we review some previous work on belief propagation and compare these results with ours. We discuss how the classic algorithm works in this case and give an example (Section 5). We then give an explanation of the reasoning behind the new algorithm as well as a presentation of the formal structure of the new algorithm (Section 6). Finally, we discuss its computational complexity (Section 7).

## 2. DEMPSTER-SHAFER THEORY

In Dempster-Shafer theory belief is assigned to a proposition by a basic probability assignment. The proposition is represented by a subset $A$ of an exhaustive set of mutually exclusive possibilities, a frame of discernment $\Theta$.

The basic probability assignment is a function from the power set of $\Theta$ to [0, 1]

$$m: 2^{\Theta} \rightarrow [0, 1]$$

whenever

$$m(\varnothing) = 0$$

and

$$\sum_{A \subseteq \Theta} m(A) = 1$$

where m($A$) is called a basic probability number, that is the belief committed exactly to $A$.

The total belief of a proposition $A$ is obtained from the sum of probabilities for those propositions that are subsets of the proposition in question and the probability committed exactly to $A$

$$\mathrm{Bel}(A) = \sum_{B \subseteq A} m(B)$$

where Bel($A$) is the total belief in $A$ and Bel(·) is called a belief function

$$\mathrm{Bel}: 2^{\Theta} \rightarrow [0, 1].$$

A subset $A$ of $\Theta$ is called a focal element of Bel if the basic probability number for $A$ is non-zero.

In addition to the belief in a proposition $A$ it is also of interest to know how plausible a proposition might be, i.e., the degree to which we do not doubt $A$. The plausibility,

$$\mathrm{Pls}: 2^{\Theta} \rightarrow [0, 1]$$

is defined as

$$\mathrm{Pls}(A) = 1 - \mathrm{Bel}(A^c).$$

We can calculate the plausibility directly from the basic probability assignment

$$\mathrm{Pls}(A) = \sum_{B \cap A \neq \emptyset} m(B).$$

Thus, while belief in A measures the total probability certainly committed to $A$, plausibility measures the total probability that is in or can be moved into $A$, i.e., $\mathrm{Bel}(A) \leq \mathrm{Pls}(A)$.

If we receive a second item of information concerning the same issue from a different source, the two items can be combined to yield a more informed view. Combining two belief functions is done by calculating the orthogonal combination with Dempster's rule. This is most simply illustrated through the combination of basic probability assignments. Let $A_i$ be a focal element of $\mathrm{Bel}_1$ and let $B_j$ be a focal element of $\mathrm{Bel}_2$. Combining the corresponding basic probability assignments $m_1$ and $m_2$ results in a new basic probability assignment $m_1 \oplus m_2$

$$m_1 \oplus m_2(A) = \mathrm{K} \cdot \sum_{A_i \cap B_j = A} m_1(A_i) \cdot m_2(B_j)$$

where K is a normalizing constant

$$\mathrm{K} = \left(1 - \sum_{A_i \cap B_j = \emptyset} m_1(A_i) \cdot m_2(B_j)\right)^{-1}.$$

This normalization is needed since, by definition, no probability mass may be committed to $\emptyset$. The new belief function $\mathrm{Bel}_1 \oplus \mathrm{Bel}_2(\cdot)$ can be calculated by the above formula from $m_1 \oplus m_2(\cdot)$.

When we wish to combine several belief functions this is simply done by combining the first two and then combine the result with the third and so forth.

## 3. DISCUSSION OF PROBLEM DOMAINS

### 3.1. Problem domains that satisfy the assumptions of the algorithm

The algorithm presented in this paper is a special case algorithm for evidences ordered in a complete directed acyclic graph, where the vertices represent states and the edges transitions between states. We are interested in finding through which sequence of states a process has developed. At every vertex we have evidence supporting the proposition that this vertex is included in the sequence and at every edge evidence expressing the degree of doubt about a transition between the corresponding states.

As an example we may consider a graph where a state represents a point in time and space and the sequence of states represents a path along which some object may have moved. For some coordinates we have evidences whose proposition tells us that this geographical point has been passed by the object at a certain time. The graph consists only of coordinates for which there is evidence. The propositions of the evidences on the edges may, for example, tell us that the time difference between the states may be to small in relation to their distance. Of course, it is impossible to move from a vertex to a previous one. There may also exist other domain-specific restrictions on the edges.

Here, we are making the assumption that only one path at a time is permitted through the graph, i.e., two objects cannot pass through the graph at the same time. The problem of analyzing paths of multiple objects can be solved by partitioning the evidences into clusters (Schubert [6]), each cluster representing a separate object, after which the problem may be solved separately for each partitioning.

The new algorithm was developed for an anti-submarine intelligence analysis system (Bergsten et al. [7]). In this application information about foreign submarine activity derives from visual observations and military sensor signals. The information is of varying quality with considerable uncertainty. Visual observations may include anything from a civilian reporting unusual wave movements on the surface to a group of naval officers recognizing a submarine tower. In shallow waters sensors may have difficulty in discerning a target and there may be several targets present simultaneously. Thus, a non-firing sensor does not necessarily exclude a passage. Weather, wind, and water temperature are other important factors determining the range and detection probability of a sensor. From this follows that an unknown number of observations may be false, i.e., not arising from submarines.

We are interested in finding the path along which the suspected submarine has moved, i.e., which observations are true. The problem we are treating here is simplified by the assumption that all observations arise from only one submarine.

This problem may be described by the complete directed acyclic graph discussed above. Each observation at a vertex, whether visual or originating from a sensor, is an evidence indicating that a submarine has visited the point of the observation. The vertices are ordered according to the time of the observations. Evidences at the edges, against transitions between the observations, appears as a lack of sensor signals, unrealistic velocity requirements, etc.

In this case we often have less than ten interesting observations during a certain period. This is because the incoming flow of observations is rather small, and observations soon become too old to give valuable information about the current position of the submarine.

Even with this moderate number of observations, the computational complexity becomes too high for the classic algorithm to be used, but is acceptable for the new algorithm.

### 3.2. Evidential Reasoning in a Complete Directed Acyclic Graph

Let a complete directed acyclic graph $G$ be given. We are interested in transitions between vertices and search for the most probable path through the graph. Every vertex $v_i$ in $G$ is associated with an evidence $e_i$ which to the amount $p_i$ supports the proposition that this vertex belongs to the sought path $S$. Furthermore, for every pair of vertices $v_i$ and $v_j$, there is an edge between the vertices that is associated with an evidence $e_{ij}$ which to the amount $q_{ij}$ speaks against a direct transition between these two vertices. Thus, $e_{ij}$ supports the proposition that there is no transition between the vertices $v_i$ and $v_j$ that does not involve any other vertex between them (Figure 1). All the corresponding belief functions are simple support functions. Since the directed acyclic graph is complete, the set of vertices is totally ordered. All evidences are supposed to be independent.

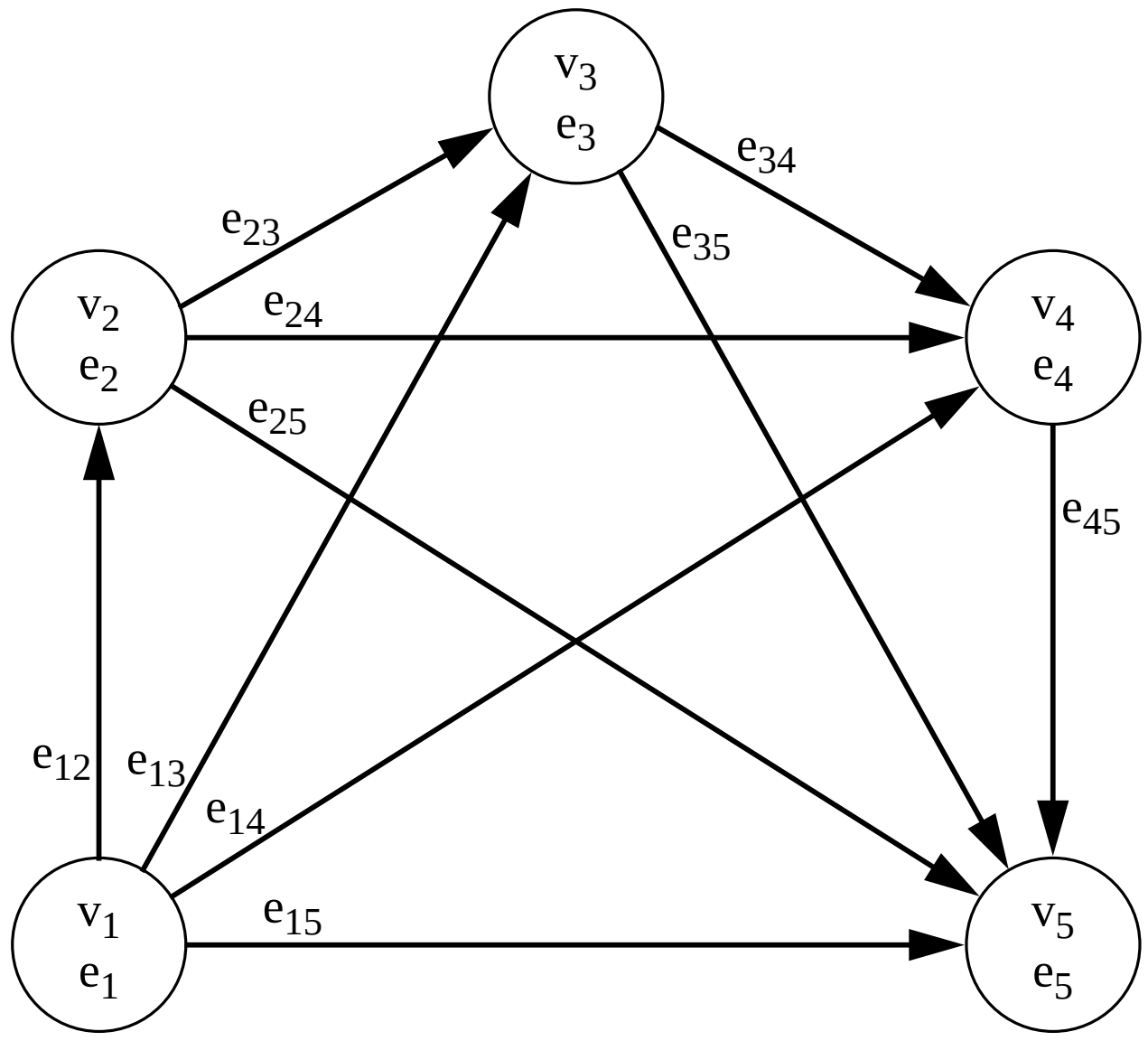

**Figure 1.** Evidences in the complete directed acyclic graph.

The first step in applying evidential reasoning to a given problem is to delimit the propositional space of possible situations, i.e., "the frame of discernment." In our case the frame of discernment is the set of all possible paths through the graph, where transitions are possible only from lower- to higher-ranked vertices. Assuming the graph $G$ consists of $n$ vertices, any path $S$ of the frame through $G$ can be represented by $<x_1, x_2, ..., x_n>$ where the i:th element corresponds to vertex $v_i$ and takes the value $r_i$ or $\neg r_i$ according to whether or not it is contained in this particular path. Our frame $\Theta$ will then consist of these $2^n$ different paths. Consider for example a graph consisting of 5 vertices $v_1, ..., v_5$ and directed edges from every vertex to all vertices with a higher index. The path from $v_1$ to $v_4$ to $v_5$, not including $v_2$ or $v_3$, is represented by $<r_1, \neg r_2, \neg r_3, r_4, r_5>$. To be able to express subsets of $\Theta$ in a convenient way, we extend the range of $x_i$ with the value $\theta_i$, meaning either $r_i$ or $\neg r_i$. E.g., $<r_1, \theta_2, \theta_3, r_4, r_5>$, an incompletely specified path, will denote all paths passing through $v_1$, $v_4$ and $v_5$.

## 4. PREVIOUS WORK

There has been some work on generally applicable improvements of the time complexity of Dempster's rule, e.g. [8,9], reducing the time complexity in the general case from $O(3^{|\Theta|})$ to $O(|\Theta| \cdot 2^{|\Theta|})$. However, most improvements have concerned important special cases. Foremost among these are methods dealing with belief propagation in trees.

### 4.1. Belief propagation in hierarchies

In 1985 Gordon and Shortliffe [10] suggested that when evidence supports singletons or disjoint subsets of the frame, a hierarchical network of subsets could be pruned to a hierarchical tree. The assumption is that a strict hierarchy of hypotheses can be defined from some subsets of $2^{\Theta}$ and that a system will only receive information for these subsets. They proposed a method partly based on the work of Barnett [11] for reasoning about hypotheses with hierarchical relationships.

Barnett showed that simple support functions focused on singletons or their complements can be combined with a time complexity, for each considered subset of $\Theta$, that is linear in the size of the frame, $|\Theta|$. In order to obtain linear time complexity, it is assumed that simple support functions with the same focus have already been combined.

Barnett's method can be described as first combining all simple support functions with equal foci and then, for each singleton, combining the resulting simple support functions for and against the singleton. For each singleton, this results in a separable support function with three focal elements: the singleton, its complement, and Θ. Finally, the separable support functions are combined separately for each considered subset of Θ in such a way that a linear time complexity is obtained. Barnett's technique will also work when the simple support functions are focused on subsets or their complements if all subsets considered are disjoint.

Besides the assumption that the domain allows a hierarchical network to be pruned to a hierarchical tree and that a system will only receive information about those subsets of the frame that are in the tree, the method by Gordon and Shortliffe is approximate in that it does not assign belief to subsets that are not in the tree. It is this approximation that changes the time complexity from exponential to linear.

The first step is borrowed from Barnett's method. All evidences with equal foci, confirming and disconfirming, are combined, with the only difference that what Barnett did with simple support functions focused on singletons is done here for all subsets of the frame that are in the tree, T. Now there are two bpa's for each subset of the frame that is in the tree, one confirming the subset and one disconfirming it; we want to combine all bpa's in the entire tree. However, combining bpa's where some focal elements are complements of subsets in the tree might produce an intersection that is not a subset or a complement of a subset that is in the tree. We begin with the confirming bpa's. These are easily combined since the intersection between two focal elements is either empty or the smaller of the two sets. This is because of the tree structure where the focal element of a child is a subset of the focal element of the parent and where focal elements at different branches are disjoint. Finally, the disconfirming bpa's are combined one by one with $m_T$, where $m_T$ is the result of the combination of all confirming bpa's. When belief is assigned to a subset, $X$, that is not in the tree this belief is reassigned to the smallest subset, $A_i$, such that $X$ is a proper subset of $A_i$, $X \subset A_i$.

Shafer and Logan [12] improved on the method by Gordon and Shortliffe. They showed that, while the algorithm by Gordon and Shortliffe usually produced a good approximation its performance was not as good when used with highly conflicting evidence. Besides not being approximate, the algorithm by Shafer and Logan also calculates belief for $A_i^c$ of every partition, $A_i$, that is in the tree, thus it calculates the plausibility for all partitions in the tree. Both algorithms run in linear time. Interestingly, Shafer and Logan showed that the linear time complexity of their algorithm is linear in the number of the nonterminal nodes due to the local computations of their algorithm and linear in the tree's branching factor due to Barnett's approach.

The algorithm by Shafer and Logan can handle evidence and calculate belief in partitions of the form $\{A_i, A_i^c\}$ for all subsets, $A_i$, in the tree. It can also calculate belief in partitions of the form $C_{A_i} \cup \{A_i^c\}$, where $C_{A_i}$ is the set of children of $A_i$. However, their algorithm can not handle evidence for $C_{A_i} \cup \{A_i^c\}$. Since these two types of evidence correspond to data and domain knowledge respectively, this is a significant restriction. A generalization of the algorithm by Shafer and Logan that manages to take domain knowledge into account is the method for belief propagation in qualitative Markov trees by Shafer, Shenoy and Mellouli [13]. In a qualitative Markov tree the children are qualitatively conditionally independent [14] given the parent, i.e., in determining which element of a child is true, there is no additional information in knowing which element of another child is true once we know which element of the parent is true. Qualitative Markov trees can arise through constructing what Shafer, Shenoy and Mellouli call the tree of families and dichotomies. This is simply done by substituting each nonterminal node with subset $A_i$ in a hierarchical tree by a parent-child pair with the dichotomy $\{A_i, A_i^c\}$ as subset at the parent and the family $C_{A_i} \cup \{A_i^c\}$ as subset at the child and furthermore substituting terminal nodes with subset $A_i$ with the dichotomy $\{A_i, A_i^c\}$.

In [15] Shenoy and Shafer list the axioms under which local computations at the nodes are possible.

Shafer, Shenoy and Mellouli point out that this computational scheme reduces the time complexity from being exponential in the size of the frame to being exponential in the size of the largest partition.

### 4.2. Comparison with our method

Barnett [11] showed that it is possible to implement Dempster's rule with a time complexity linear in the size of the frame, $|\Theta|$, when the belief functions being combined are all simple support functions focused on singletons or their complements. In our case, however, the simple support functions are never focused on singletons and, with one exception, not focused on the complements of singletons. Our frame consists of all possible single paths in a complete directed acyclic graph, and the simple support functions are on subsets representing individual vertices in the complete directed acyclic graph or on subsets representing the direct transition between two vertices, i.e., on elements of $2^\Theta$ that are not singletons or, with the exception of the two vertex graph, their complements.

Gordon and Shortliffe suggest that when evidences support singletons or disjoint subsets of the frame the hierarchical network of subsets could be pruned to a tree. Then they suggested methods for the combination of evidence in trees. Our case can of course also be represented with a hierarchical network of subsets, as seen by the example in Figure 2 of a hierarchical network of subsets of the frame for a two-vertex graph. As mentioned above, we never have evidence supporting singletons and the subsets of the frame that are supported are not disjoint. In Figure 2 the last subset of the second row and the first two subsets of the third row are supported by one simple support function each. This is support offered against the belief in both vertices, i.e., support offered for the complement of the belief in both vertices, $<r_1, r_2>^c$, support offered for the first vertex and support offered for the second vertex respectively. Because the supported subsets in the hierarchical network of our problem are not disjoint, we can not prune our network to a tree and use the scheme suggested by Gordon and Shortliffe.

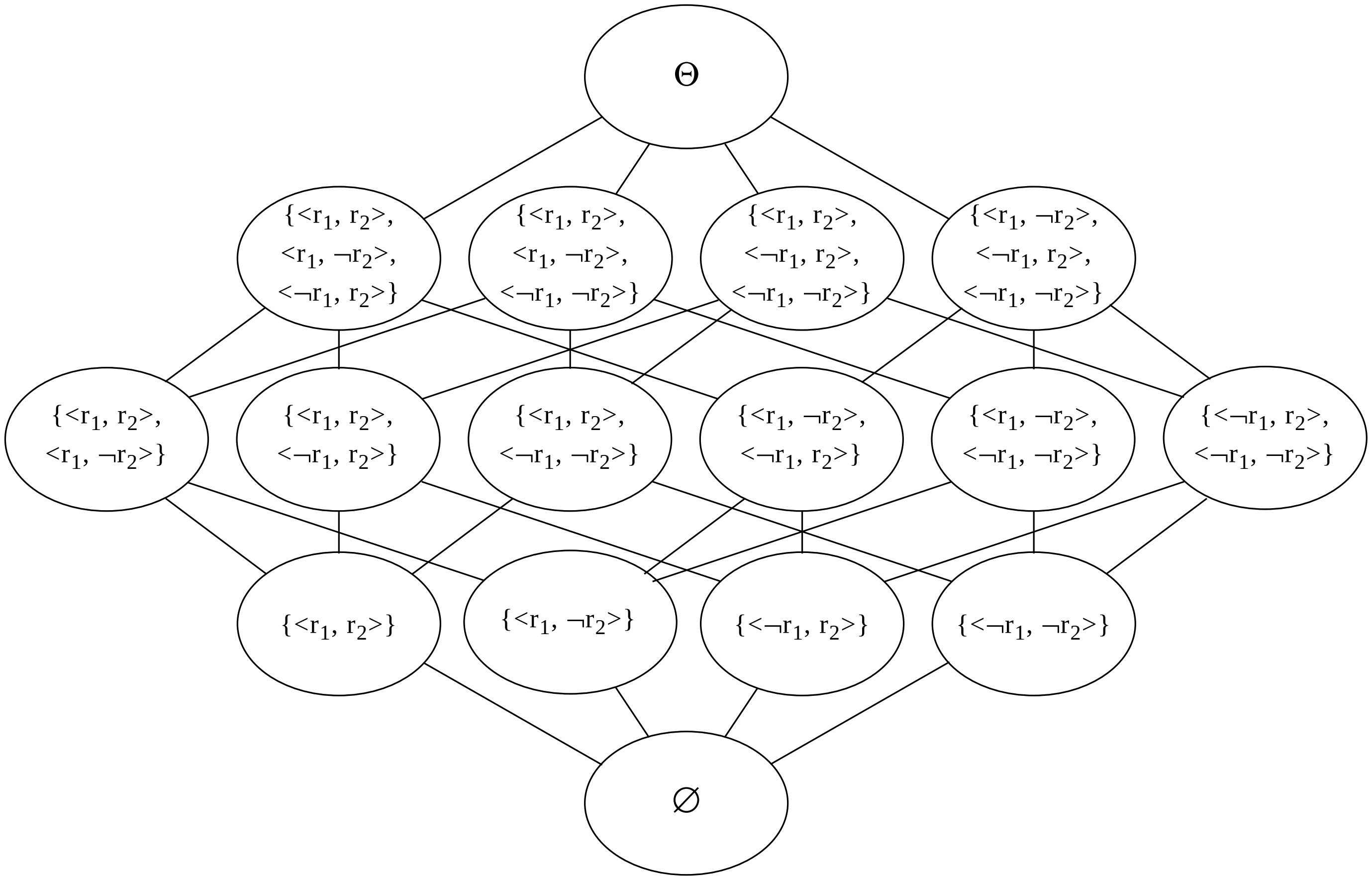

**Figure 2.** Hierarchical network of subsets for a two-vertex graph.

The two other papers by Shafer and Logan [12] and Shafer, Shenoy and Mellouli [13] concern the case of belief propagation in qualitative Markov trees only. Thus, the methods presented in these three papers are not applicable in the case with evidences in a non-prunable network of subsets.

Instead of propagating the belief in a hierarchical structure of subsets our algorithm reasons, separately for each instance of the frame, about the logical conditions of the completely specified path through the complete directed acyclic graph.

## 5. DEMPSTER'S RULE - THE CLASSIC ALGORITHM

Let us for convenience define the representation of a path as a conjunction of $n$ propositions,

$$<x_1, x_2, \ldots, x_n> \triangleq x_1 \wedge x_2 \wedge \ldots \wedge x_n,$$

and define

$$\langle x_1, x_2, \ldots, x_n \rangle \triangleq x_1 \vee x_2 \vee \ldots \vee x_n$$

as a disjunction of $n$ paths. We have

$$<x_1, x_2, \ldots, x_n> \wedge <y_1, y_2, \ldots, y_n> = <x_1 \wedge y_1, x_2 \wedge y_2, \ldots, x_n \wedge y_n>.$$

and

$$\langle x_1, x_2, \ldots, x_n \rangle \wedge \langle y_1, y_2, \ldots, y_n \rangle$$
$$= \langle x_1 \wedge y_1, x_1 \wedge y_2, \ldots, x_1 \wedge y_n, x_2 \wedge y_1, x_2 \wedge y_2, \ldots, x_2 \wedge y_n, \ldots,$$
$$x_n \wedge y_1, x_n \wedge y_2, \ldots, x_n \wedge y_n \rangle.$$

In the problem of transfers between vertices in a graph, where a transfer might be possible only from a vertex with lower index to a vertex with higher index, our focus is on paths that may consist of several vertices. The frame of discernment is the set of all completely specified paths, $\Theta = \{<x_1, x_2, \ldots, x_n> | \forall i.x_i \in \{r_i, \neg r_i\}\}$, where $r_i$ is the proposition of the evidence corresponding to vertex $i$ in the graph, $x_i = r_i$ means that the vertex $v_i$ is included in the path and $x_i = \neg r_i$ means that $v_i$ is not included in the path.

We are interested in the problem where one begins with a basic probability assignment for those elements that belongs to the following subset of $2^{\Theta}$:

$$\{<\theta_1, \ldots, \theta_{i-1}, r_i, \theta_{i+1}, \ldots, \theta_n>\}$$
$$\cup \{\langle <\theta_1, \ldots, \theta_{i-1}, \neg r_i, \theta_{i+1}, \ldots, \theta_n>,$$
$$<\theta_1, \ldots, \theta_{j-1}, \neg r_j, \theta_{j+1}, \ldots, \theta_n>,$$
$$\langle \{<\theta_1, \ldots, \theta_{k-1}, r_k, \theta_{k+1}, \ldots, \theta_n> | \forall k.i<k<j\}\rangle\rangle | \forall i,j.i<j\},$$

that is, we begin with positive evidence, $e_i$, for all vertices and negative evidence, $e_{ij}$, against all directed edges $v_i$ to $v_j$ where $i < j$–evidence that the path does not include $v_i$ or that it does not include $v_j$ or that it does include a vertex $v_k$, $i < k < j$, between $v_i$ and $v_j$, thus excluding any direct transfer.

Thus, we have the following two types of evidences to consider.

**1.** The evidence $e_i$ for every vertex in the graph. The bpa for the path with a single evidence $e_i$ is

$$m_i(<x_1, x_2, \ldots, x_n>) = \begin{cases} p_i, & (x_i = r_i) \wedge (\forall k | k \neq i.x_k = \theta_k) \\ 1-p_i, & \forall k.x_k = \theta_k \\ 0, & \text{otherwise} \end{cases},$$

**2.** The evidence $e_{ij}$ against the edges between every two vertices in the graph. The corresponding bpa is here

$$m_{ij}\left(\begin{matrix}\langle <x_1, \ldots, x_i, \ldots, x_n>, <x_1, \ldots, x_j, \ldots, x_n>, \\ \langle \{<x_1, \ldots, x_k, \ldots, x_n> | \forall k.i<k<j\}\rangle\rangle\end{matrix}\right)$$
$$= \begin{cases} q_{ij}, & ((x_i = \neg r_i) \wedge (\forall s \neq i.x_s = \theta_s)) \vee \\ & ((x_j = \neg r_j) \wedge (\forall s \neq j.x_s = \theta_s)) \vee \\ & ((x_k = r_k) \wedge (\forall s.x_s = \theta_s)) \\ 1-q_{ij}, & \forall s.x_s = \theta_s \\ 0, & \text{otherwise} \end{cases},$$

All these evidences are to be combined using Dempster's rule. The evidences can be combined in an arbitrary order because Dempster's rule is both associative and commutative.

*correction inserted*

### 5.1. Explaining the classic algorithm

We will seek the support and plausibility for all elements of $\Theta$ that are of the form $<x_1, x_2, \ldots, x_n>$ where $x_i \in \{r_i, \neg r_i\}$. For the sake of simplicity we shall first use Dempster's rule to separately fuse all positive evidences and all negative evidences,

$$\mathrm{Bel}_p = \oplus \{\mathrm{Bel}_{<\theta_1, \ldots, \theta_{i-1}, r_i, \theta_{i+1}, \ldots, \theta_n>}\},$$

$$\mathrm{Bel}_n = \oplus \{\mathrm{Bel}_{\langle <\theta_1, \ldots, \theta_{i-1}, \neg r_i, \theta_{i+1}, \ldots, \theta_n>, <\theta_1, \ldots, \theta_{j-1}, \neg r_j, \theta_{j+1}, \ldots, \theta_n>, \langle \{<\theta_1, \ldots, \theta_{k-1}, r_k, \theta_{k+1}, \ldots, \theta_n> | \forall k. i < k < j\} \rangle \rangle} | \forall i, j. i < j\},$$

thus leaving the conflict creating fusion, $\mathrm{Bel}_p \oplus \mathrm{Bel}_n$, until last. The first of these fusions are shown in Figure 3 and Figure 4. The support and plausibility of all paths will then be calculated from the result of the last fusion.

In a fusion of two belief functions the representation in every intersection of focal elements is the conjunction on these focal elements' representations. The value of that intersection is the product of the values of the focal elements. In the upper left quadrant of Figure 4, for example, the result is derived from:

$$\begin{aligned}
&\langle <\neg r_1, \theta_2, \ldots, \theta_n>, <\theta_1, \neg r_2, \theta_3, \ldots, \theta_n> \rangle \\
&\wedge \langle <\neg r_1, \theta_2, \ldots, \theta_n>, <\theta_1, \theta_2, \neg r_3, \theta_4, \ldots, \theta_n>, \\
&\quad <\theta_1, r_2, \theta_3, \ldots, \theta_n> \rangle \\
&\quad = \langle <\neg r_1, \theta_2, \ldots, \theta_n>, <\neg r_1, \theta_2, \neg r_3, \theta_4, \ldots, \theta_n>, \\
&\qquad <\neg r_1, r_2, \theta_3, \ldots, \theta_n>, <\neg r_1, \neg r_2, \theta_3, \ldots, \theta_n>, \\
&\qquad <\theta_1, \neg r_2, \neg r_3, \theta_4, \ldots, \theta_n>, \emptyset \rangle \\
&\quad = \{\text{since the second, third and fourth} \\
&\qquad\quad \text{elements are contained in the first}\} \\
&\quad = \langle <\neg r_1, \theta_2, \ldots, \theta_n>, <\theta_1, \neg r_2, \neg r_3, \theta_4, \ldots, \theta_n> \rangle
\end{aligned}$$

| | $<\theta_1, r_2, \theta_3, \ldots, \theta_n>$ $p_2$ | $<\theta_1, \ldots, \theta_n>$ $1 - p_2$ |
|---|---|---|
| $<r_1, \theta_2, \ldots, \theta_n>$ $p_1$ | $<r_1, r_2, \theta_3, \ldots, \theta_n>$ $p_1 \cdot p_2$ | $<r_1, \theta_2, \ldots, \theta_n>$ $p_1 \cdot (1 - p_2)$ |
| $<\theta_1, \ldots, \theta_n>$ $1 - p_1$ | $<\theta_1, r_2, \theta_3, \ldots, \theta_n>$ $(1 - p_1) \cdot p_2$ | $<\theta_1, \ldots, \theta_n>$ $(1 - p_1) \cdot (1 - p_2)$ |

**Figure 3.** The first use of Dempster's rule on positive evidence.

*correction inserted*

| | $\langle <\neg r_1, \theta_2, ..., \theta_n>,$<br>$<\theta_1, \theta_2, \neg r_3, \theta_4, ..., \theta_n>,$<br>$<\theta_1, r_2, \theta_3, ..., \theta_n>\rangle$<br>$q_{13}$ | $<\theta_1, ..., \theta_n>$<br>$1 - q_{13}$ |
|---|---|---|
| $\langle <\neg r_1, \theta_2, ..., \theta_n>,$<br>$<\theta_1, \neg r_2, \theta_3, ..., \theta_n>\rangle$<br>$q_{12}$ | $\langle <\neg r_1, \theta_2, ..., \theta_n>,$<br>$<\theta_1, \neg r_2, \neg r_3, \theta_4, ..., \theta_n>\rangle$<br>$q_{12} \cdot q_{13}$ | $\langle <\neg r_1, \theta_2, ..., \theta_n>,$<br>$<\theta_1, \neg r_2, \theta_3, ..., \theta_n>\rangle$<br>$q_{12} \cdot (1 - q_{13})$ |
| $<\theta_1, ..., \theta_n>$<br>$1 - q_{12}$ | $\langle <\neg r_1, \theta_2, ..., \theta_n>,$<br>$<\theta_1, \theta_2, \neg r_3, \theta_4, ..., \theta_n>,$<br>$<\theta_1, r_2, \theta_3, ..., \theta_n>\rangle$<br>$(1 - q_{12}) \cdot q_{13}$ | $<\theta_1, ..., \theta_n>$<br>$(1 - q_{12}) \cdot (1 - q_{13})$ |

**Figure 4.** The first use of Dempster's rule on negative evidence.

and

$$m(\langle <\neg r_1, \theta_2, \ldots, \theta_n>, <\theta_1, \neg r_2, \theta_3, \ldots, \theta_n>\rangle) \cdot m(\langle <\neg r_1, \theta_2, \ldots, \theta_n>, <\theta_1, \theta_2, \neg r_3, \theta_4, \ldots, \theta_n>, <\theta_1, r_2, \theta_3, \ldots, \theta_n>\rangle) = q_{12} \cdot q_{13}.$$

The fusion, Figure 4, will result in a new basic probability assignment with basic probability numbers for all new representations. The basic probability number of $\langle <\neg r_1, \theta_2, \ldots, \theta_n>, <\theta_1, \neg r_2, \neg r_3, \theta_4, \ldots, \theta_n>\rangle$ for instance, is the normalized sum of values from all intersections with exactly this representation. In Figure 4 there are, of course, no other intersections with this representation and no conflict to cause a normalization. A new belief function is given by the new basic probability assignment and the belief of a proposition, $A$, is the sum of the basic probability numbers for that proposition, m($A$), and all propositions that are proper subsets of $A$, $m(B \mid B \subset A)$. In our case, however, the situation is somewhat simpler because we are only seeking the support and plausibility of propositions that have no proper subsets.

Let us, for simplicity, observe the final fusion $\mathrm{Bel}_p \oplus \mathrm{Bel}_n$ in the case with three vertices, Figure 5. In each square the representation and its value is derived in the same way as above. The support and plausibility of all elements can be calculated as:

$$\forall x_i \big| x_i \in \{r_i, \neg r_i\} \,.\mathrm{Spt}(<x_1, x_2, \ldots, x_n>) = \frac{1}{1-k} \sum_{A_i \cap A_j = <x_1, x_2, \ldots, x_n>} m(A_i) \cdot m(A_j),$$

$$\forall x_i \big| x_i \in \{r_i, \neg r_i\} \,.\mathrm{Pls}(<x_1, x_2, \ldots, x_n>) = \frac{1}{1-k} \sum_{<x_1, x_2, \ldots, x_n> \in A_i \cap A_j} m(A_i) \cdot m(A_j)$$

where

$$k = \sum_{A_i \cap A_j = \emptyset} m(A_i) \cdot m(A_j),$$

$A_i, A_j \subset 2^{\Theta}$ are focal elements in the last fusion and

$$<x_1, x_2, \ldots, x_n> \in \langle y_1, y_2, \ldots, y_n \rangle \quad \text{iff } \exists y_i \big| <x_1, x_2, \ldots, x_n> \in y_i,$$
$$<x_1, x_2, \ldots, x_n> \in <z_1, z_2, \ldots, z_n> \quad \text{iff } \forall i.\,(z_i = x_i) \vee (z_i = \theta_i)\,.$$

Due to the high computational complexity it is only possible to perform these computations for graphs consisting of very few vertices. This problem is solved by a new algorithm, where instead of performing all combinations step-by-step, the final result is derived directly by reasoning about the completely specified paths from the beginning.

### 5.2. An example

Consider the path $<r_1, \neg r_2, r_3>$. Before we calculate the support and plausibility of the path we must calculate the conflict, $k$, in the final fusion $\mathrm{Bel}_p \oplus \mathrm{Bel}_n$. The conflict is the sum of all contribution from all intersections $A_i \cap A_j = \emptyset$, Figure 5;

| | $<r_1, r_2, r_3>$ $p_1 \cdot p_2 \cdot p_3$ | $<\theta_1, r_2, r_3>$ $(1 - p_1) \cdot p_2 \cdot p_3$ | $<r_1, \theta_2, r_3>$ $p_1 \cdot (1 - p_2) \cdot p_3$ | $<r_1, r_2, \theta_3>$ $p_1 \cdot p_2 \cdot (1 - p_3)$ |
|---|---|---|---|---|
| $\langle <\neg r_1, \neg r_2, \theta_3>, <\neg r_1, \theta_2, \neg r_3>, <\theta_1, \neg r_2, \neg r_3> \rangle$ $q_{12} \cdot q_{13} \cdot q_{23}$ | ∅ | ∅ | ∅ | ∅ |
| $\langle <\neg r_1, \neg r_2, \theta_3>, <\theta_1, \theta_2, \neg r_3> \rangle$ $(1 - q_{12}) \cdot q_{13} \cdot q_{23}$ | ∅ | ∅ | ∅ | $<r_1, r_2, \neg r_3>$ |
| $\langle <\neg r_1, \theta_2, \neg r_3>, <\theta_1, \neg r_2, \theta_3> \rangle$ $q_{12} \cdot (1 - q_{13}) \cdot q_{23}$ | ∅ | ∅ | $<r_1, \neg r_2, r_3>$ | ∅ |
| $\langle <\neg r_1, \theta_2, \theta_3>, <\theta_1, \neg r_2, \neg r_3> \rangle$ $q_{12} \cdot q_{13} \cdot (1 - q_{23})$ | ∅ | $<\neg r_1, r_2, r_3>$ | ∅ | ∅ |
| $\langle <\theta_1, \neg r_2, \theta_3>, <\theta_1, \theta_2, \neg r_3> \rangle$ $(1 - q_{12}) \cdot (1 - q_{13}) \cdot q_{23}$ | ∅ | ∅ | $<r_1, \neg r_2, r_3>$ | $<r_1, r_2, \neg r_3>$ |
| $\langle <\neg r_1, \theta_2, \theta_3>, <\theta_1, \theta_2, \neg r_3>, <\theta_1, r_2, \theta_3> \rangle$ $(1 - q_{12}) \cdot q_{13} \cdot (1 - q_{23})$ | $<r_1, r_2, r_3>$ | $<\theta_1, r_2, r_3>$ | $<r_1, r_2, r_3>$ | $<r_1, r_2, \theta_3>$ |
| $\langle <\neg r_1, \theta_2, \theta_3>, <\theta_1, \neg r_2, \theta_3> \rangle$ $q_{12} \cdot (1 - q_{13}) \cdot (1 - q_{23})$ | ∅ | $<\neg r_1, r_2, r_3>$ | $<r_1, \neg r_2, r_3>$ | ∅ |
| $<\theta_1, \theta_2, \theta_3>$ $(1 - q_{12}) \cdot (1 - q_{13}) \cdot (1 - q_{23})$ | $<r_1, r_2, r_3>$ | $<\theta_1, r_2, r_3>$ | $<r_1, \theta_2, r_3>$ | $<r_1, r_2, \theta_3>$ |

**Figure 5.** The last use of Dempster's rule: fusion positive and negative evidence.

$$k = \ldots = p_1 \cdot p_2 \cdot q_{12} + p_1 \cdot (1 - p_2) \cdot p_3 \cdot (q_{12} \cdot q_{13} + q_{13} \cdot q_{23} - q_{12} \cdot q_{13} \cdot q_{23}) + p_2 \cdot p_3 \cdot q_{23} - p_1 \cdot p_2 \cdot p_3 \cdot q_{12} \cdot q_{23}.$$

| $<\theta_1, \theta_2, r_3>$ $(1-p_1)\cdot(1-p_2)\cdot p_3$ | $<\theta_1, r_2, \theta_3>$ $(1-p_1)\cdot p_2\cdot(1-p_3)$ | $<r_1, \theta_2, \theta_3>$ $p_1\cdot(1-p_2)\cdot(1-p_3)$ | $<\theta_1, \theta_2, \theta_3>$ $(1-p_1)\cdot(1-p_2)\cdot(1-p_3)$ |
|---|---|---|---|
| $<\neg r_1, \neg r_2, r_3>$ | $<\neg r_1, r_2, \neg r_3>$ | $<r_1, \neg r_2, \neg r_3>$ | $\langle <\neg r_1, \neg r_2, \theta_3>, <\neg r_1, \theta_2, \neg r_3>, <\theta_1, \neg r_2, \neg r_3>\rangle$ |
| $<\neg r_1, \neg r_2, r_3>$ | $<\theta_1, r_2, \neg r_3>$ | $<r_1, \theta_2, \neg r_3>$ | $\langle <\neg r_1, \neg r_2, \theta_3>, <\theta_1, \theta_2, \neg r_3>\rangle$ |
| $<\theta_1, \neg r_2, r_3>$ | $<\neg r_1, r_2, \neg r_3>$ | $<r_1, \neg r_2, \theta_3>$ | $\langle <\neg r_1, \theta_2, \neg r_3>, <\theta_1, \neg r_2, \theta_3>\rangle$ |
| $<\neg r_1, \theta_2, r_3>$ | $<\neg r_1, r_2, \theta_3>$ | $<r_1, \neg r_2, \neg r_3>$ | $\langle <\neg r_1, \theta_2, \theta_3>, <\theta_1, \neg r_2, \neg r_3>\rangle$ |
| $<\theta_1, \neg r_2, r_3>$ | $<\theta_1, r_2, \neg r_3>$ | $\langle <r_1, \neg r_2, \theta_3>, <r_1, \theta_2, \neg r_3>\rangle$ | $\langle <\theta_1, \neg r_2, \theta_3>, <\theta_1, \theta_2, \neg r_3>\rangle$ |
| $\langle <\neg r_1, \theta_2, r_3>, <\theta_1, r_2, r_3>\rangle$ | $<\theta_1, r_2, \theta_3>$ | $\langle <r_1, \theta_2, \neg r_3>, <r_1, r_2, \theta_3>\rangle$ | $\langle <\neg r_1, \theta_2, \theta_3>, <\theta_1, \theta_2, \neg r_3>, <\theta_1, r_2, \theta_3>\rangle$ |
| $\langle <\neg r_1, \theta_2, r_3>, <\theta_1, \neg r_2, r_3>\rangle$ | $<\neg r_1, r_2, \theta_3>$ | $<r_1, \neg r_2, \theta_3>$ | $\langle <\neg r_1, \theta_2, \theta_3>, <\theta_1, \neg r_2, \theta_3>\rangle$ |
| $<\theta_1, \theta_2, r_3>$ | $<\theta_1, r_2, \theta_3>$ | $<r_1, \theta_2, \theta_3>$ | $<\theta_1, \theta_2, \theta_3>$ |

**Figure 5.** *Continued.*

The support is calculated as the normalized sum of all contributions from the intersection whose representation is identical with the path. Thus the support of $<r_1, \neg r_2, r_3>$ is the contributions, in Figure 5, from row 3 column 3, row 5 column 3 and row 7 column 3;

$$\text{Spt}(<r_1, \neg r_2, r_3>) = \frac{1}{1-k}(p_1 \cdot (1-p_2) \cdot p_3 \cdot q_{12} \cdot (1-q_{13}) \cdot q_{23}$$
$$+ p_1 \cdot (1-p_2) \cdot p_3 \cdot (1-q_{12}) \cdot (1-q_{13}) \cdot q_{23}$$

$$+ p_1 \cdot (1-p_2) \cdot p_3 \cdot q_{12} \cdot (1-q_{13}) \cdot (1-q_{23}) )$$

$$= \frac{1}{1-k} (p_1 \cdot (1-p_2) \cdot p_3 \cdot (1-q_{13})$$

$$\cdot (q_{12} + (1-q_{12}) \cdot q_{23}) ) .$$

When calculating the plausibility we normalize the sum of all contributions from the intersections in which representations the path is contained. These are the 16 intersections of rows 3, 5, 7, 8 and columns 3, 5, 7, 8, Figure 5.

Take for instance the intersection in row 5 column 7:

$$<r_1, \neg r_2, r_3> \in \langle <r_1, \neg r_2, \theta_3>, <r_1, \theta_2, \neg r_3> \rangle$$

since

$$<r_1, \neg r_2, r_3> \in y_1 \quad (= <r_1, \neg r_2, \theta_3>)$$

which is true since

$$\begin{cases} z_1 = x_1 \,(= r_1) \\ z_2 = x_2 \,(= \neg r_2) \\ z_3 = \theta_3. \end{cases}$$

Thus, $< r_1, \neg r_2, r_3 >$ is contained in $\langle < r_1, \neg r_2, \theta_3 >, < r_1, \theta_2, \neg r_3 > \rangle$ and the value of the intersection in row 5 column 7 is contributing to the plausibility of $<r_1, \neg r_2, r_3>$. The plausibility becomes, after some simplification:

$$\text{Pls}(<r_1, \neg r_2, r_3>) = \dots = \frac{1}{1-k} (1-p_2) \cdot (1-q_{13}) .$$

## 6. DEMPSTER'S RULE - THE NEW ALGORITHM

We are now ready to give an intuitive presentation of our algorithms for calculating the support and plausibility for all elements, $A$, of $2^\Theta$ that are of the form $< x_1, x_2, ..., x_n>$ where $x_i \in \{ r_i, \neg r_i\}$, i.e., the completely specified paths of $\Theta$. The new algorithm is built on an expression for the final result of support and plausibility, i.e., we only have to evaluate this expression instead of all stepwise combinations. The algorithms, when used symbolically, calculate the symbolic structure of the support and plausibility for a path derived through summation of intersections in the final fusion, $\text{Bel}_p \oplus \text{Bel}_n$, of the classic algorithm. This means that the new algorithm can comparatively quickly calculate the answers which had to be calculated through a lot of time-consuming fusions and pattern-matching summations in the classic algorithm. We will first explain the mathematical reasoning behind this algorithm, which calculates support and plausibility in the following steps: unnormalized plausibility, unnormalized support, conflict and finally plausibility and support normalized by the conflict.

### 6.1. Plausibility

Let us start with the plausibility and see what is sufficient to make a path plausible. Plausibility for a path means to which degree this path is possible, i.e., to which degree no known factors speak against this path. There are only two types of items which speak against a path–the positive evidence for vertices that are not included in the path and the negative evidence against edges between vertices that are included in the path. This means that the degree to which we do not assign support to these evidences equals the degree to which the path is possible. The algorithm for plausibility is then

$$\mathrm{Pls}(S) = \mathrm{Pls}^*(S) / (1 - k)$$

where $k$ is the conflict and $\mathrm{Pls}^*(\cdot)$ is the unnormalized plausibility

$$\mathrm{Pls}^*(S) = \prod_{\forall i | v_i \notin S} (1 - p_i) \cdot \prod_{\forall i} (1 - q_{s_i, s_{i+1}}),$$

where $q_{ij}$ is the degree of doubt of the edge between vertices $v_i$ and $v_j$ and $v_{s_i}$ is the $i$:th vertex in the path $S$.

### 6.2. Support

The algorithm for support is much more complicated than the one for plausibility. It is not only necessary to find out which evidence speaks against the path; it is also necessary to insist that the evidence of the vertices and edges that are included in the path speaks in favor of it.

While each of the evidences supports only one proper subset of Θ, i.e., corresponding belief functions are simple support functions, we will say for the sake of simplicity that the evidence $e_i$ is false (true) when we mean that the proposition according to the proper subset is false (true). The same holds for the evidences $e_{ij}$.

6.2.1. EXPLAINING THE ALGORITHM FOR SUPPORT Assuming the path includes $m$ vertices, we first realize that the following three statements have to be true:

**1.** Every vertex in the path has to be visited.

**(a)** For the first and the last vertices in the path we only have one possibility: the evidences $e_{s_1}$ and $e_{s_m}$ are true and the support for this is $p_{s_1} \cdot p_{s_m}$, for $m > 1$ and $p_{s_1}$, $m = 1$.

**(b)** For every intermediate vertex $v_{s_i}$ in $S$ there are two different possibilities:

**(i)** The evidence $e_{s_i}$ is true. The support for this is $p_{s_i}$.

**(ii)** The evidence $e_{s_i}$ may be false, but the evidence against edges are speaking against all other ways from the last vertex visited before $v_{s_i}$, to the first vertex visited after $v_{s_i}$. The possibility that $e_{s_i}$ may be false is (1 - $p_{s_i}$).

**2.** The transitions between consecutive vertices in the path are possible, i.e., the evidence against those edges has to be false. The possibility for this is

$$\prod_{i=1}^{m-1} (1 - q_{s_i, s_{i+1}}) .$$

**3.** No vertex outside the path is permitted to be visited. We first state that the evidences $e_i$ for vertices outside the path have to be false. The possibility for this is

$$\prod_{\forall i | v_i \notin S} (1 - p_i) .$$

But even if these evidences may be false, we can not be sure that a vertex outside $S$ is not visited. In order to guarantee this we also make the following three statements:

**(a)** No transition is possible from vertices before $v_{s_1}$ to this vertex, i.e., all evidences against edges from vertices before $v_{s_1}$ to $v_{s_1}$ are true. The support for this is:

$$\prod_{i < s_1} q_{i, s_1} .$$

This statement assures that we enter the path at $v_{s_1}$.

**(b)** No transition is possible from $v_{s_m}$ to vertices after this vertex. This is to assure that $v_{s_m}$ is the last vertex in the path. The support for this is:

$$\prod_{i > s_m} q_{s_m, i} .$$

**(c)** For the vertices not belonging to $S$ which are located between vertices in $S$ we state that no transition is possible to these vertices from vertices in $S$, or if such a transition is possible then it is not possible to join the next vertex in the path we have stated to be true.

The support is now received by multiplying the contributions from the three statements. Let us illustrate this with an example.

Let the graph $G$ consist of 5 vertices $v_1$, ..., $v_5$. We shall compute the support and plausibility for the path $< r_1, \neg r_2, r_3, r_4, \neg r_5>$. The unnormalized plausibility is easily derived in the way described above:

$$\text{Pls*}( < r_1, \neg r_2, r_3, r_4, \neg r_5 > ) = (1 - p_2) \cdot (1 - p_5) \cdot (1 - q_{13}) \cdot (1 - q_{34}).$$

When computing the support we apply the three statements above. From statement (1a) we get the factor $p_1 \cdot p_4$. Considering statement (1b) forces us to break down the calculations into two parts:

**(i)** We state $e_3$ to be true.

**(ii)** We do not state $e_3$ to be true.

The factor calculated in (3c), in order to prevent visiting a vertex between the first and last vertices of the path which does not belong to the path, will differ depending on which vertices we have stated to be true, therefore we calculate the factors from the other statements separately for (i) and (ii) and then sum up the two contributions.

We begin with (i).

When $e_3$ is true, the factor from this statement is $p_3$. From (2) we get the factor $(1 - q_{13}) \cdot (1 - q_{34})$. Statement (3) states that the evidences $e_2$ and $e_5$ have to be false, giving us the factor $(1 - p_2) \cdot (1 - p_5)$. Statement (3a) can be disregarded while the first vertex in the path is the first vertex in the graph and from (3b) we get the factor $q_{45}$.

Let us now regard (3c) which states either that a transition from $v_1$ to $v_2$ is not allowed, which gives us the factor $q_{12}$, or that if a transition from $v_1$ to $v_2$ is allowed, then it must be impossible to reach the next vertex in the path stated to be true, which according to our assumption (1b) is $v_3$. This gives us the factor $(1 - q_{12}) \cdot q_{23}$, i.e., the total factor from (3c) is $q_{12} + (1 - q_{12}) \cdot q_{23}$. We have now calculated the first term of the support $p_1 \cdot p_4 \cdot p_3 \cdot (1 - q_{13}) \cdot (1 - q_{34}) \cdot (1 - p_2) \cdot (1 - p_5) \cdot q_{45} \cdot (q_{12} + (1 - q_{12}) \cdot q_{23})$.

Let us calculate the second term, (ii), where we do not state $e_3$ to be true.

The possibility for this is $(1 - p_3)$. The factors (1a), (2) and (3a-b) in this term are the same as in the term above and (3c) is in this case implied in (1b), hence it is enough to calculate (1b). We have the following two possibilities:

**(1)** transition from $v_1$ to $v_2$ or $v_4$ is impossible, which implies that the only path from $v_1$ to $v_4$ is $v_1 - v_3 - v_4$. This gives us the factor $q_{12} \cdot q_{14}$.

**(2)** transition from $v_1$ to $v_4$ is impossible but we allow a transition from $v_1$ to $v_2$ but not from $v_2$ to $v_3$ or $v_4$, giving us the factor $(1 - q_{12}) \cdot q_{14} \cdot q_{23} \cdot q_{24}$.

The second term for the support is then

$$p_1 \cdot p_4 \cdot (1 - p_3) \cdot (1 - q_{13}) \cdot (1 - q_{34}) \cdot q_{45} \cdot (1 - p_2) \cdot (1 - p_5) \\ \cdot (q_{12} \cdot q_{14} + (1 - q_{12}) \cdot q_{14} \cdot q_{23} \cdot q_{24})$$

and we end up with the unnormalized support

$$\mathrm{Sup}^*(< r_1, \neg r_2, r_3, r_4, \neg r_5>)$$
$$= p_1 \cdot (1 - p_2) \cdot p_4 \cdot (1 - p_5) \cdot (1 - q_{13})$$
$$\cdot (1 - q_{34}) \cdot q_{45} \cdot (p_3 \cdot (q_{12} + (1 - q_{12}) \cdot q_{23}) + (1 - p_3) \cdot q_{14}$$
$$\cdot (q_{12} + (1 - q_{12}) \cdot q_{23} \cdot q_{24})).$$

The normalized support becomes

$$\mathrm{Sup}(< r_1, \neg r_2, r_3, r_4, \neg r_5>) = \mathrm{Sup}^*(< r_1, \neg r_2, r_3, r_4, \neg r_5>) / (1 - k)$$

where $k$ is the conflict.

In Section 6.2.2 we present a detailed analysis of the algorithm for support, followed in Section 6.2.3 by the algorithm itself. The reader may skip these sections on a first reading and continue with Section 6.3 on conflict.

6.2.2. A DETAILED ANALYSIS OF THE ALGORITHM FOR SUPPORT First some useful definitions:

$$m^-(\omega, i) \triangleq \min(i) \mid \omega_i = r_i,\ 1 \le i \le n,$$
$$m^-(\omega, i, j) \triangleq \min(i) \mid \omega_i = r_i,\ 1 \le i, j \le n,\ i > j$$
$$m^+(\omega, i) \triangleq \max(i) \mid \omega_i = r_i,\ 1 \le i \le n,$$
$$m^+(\omega, i, j) \triangleq \max(i) \mid \omega_i = r_i,\ 1 \le i, j \le n.\ i < j.$$

Thus, $m^-(x, i)$ is the first vertex in the path, $v_{s_1}$, and $m^-(x, i, j)$ is the first vertex in the path of those with index larger than $j$.

The algorithm can be broken down into three different parts.

For the first part we have the same argument as with the plausibility, i.e., the evidence which speaks against the path must be false, thus the same terms as in the plausibility.

The second part of the algorithm is to assure that the first and last vertices of the path actually are the first and last vertices included in the path, i.e., that there is evidence against edges to the first vertex of the path from any vertices in the graph before the path’s first vertex, that the path’s first and last vertices are included in the path, and that there is evidence against edges from the last vertex of the path to any vertices in the graph after the path’s last vertex. This gives us the terms:

$$\left( \prod_{\forall j \mid 1 \le j < m^-(x, i)} q_{j, m^-(x, i)} \right) \cdot p_{m^-(x, i)}$$

and

$$\begin{cases} p_{m^+(x, i)}, & m^-(x, i) \ne m^+(x, i) \\ 1, & m^-(x, i) = m^+(x, i) \end{cases} \cdot \left( \prod_{\forall j \mid m^+(x, i) < j \le n} q_{m^+(x, i), j} \right).$$

The third part of the algorithm concerns the transfers from the first vertex of the path until the last one. The positive evidence of every internal vertex of the path, i.e., the vertices $x_j = r_j$ where $m^-(x, i) < j < m^+(x, i)$, is to some degree committed in favor of the path and is for the remainder uncommitted. However, for each combination of statements for the internal vertices, i.e., internal vertices stated or not stated to be true for the combination, we have support for the path given the right conditions for the edges. Thus, we have to sum up the contribution from all the combinations;

$$\forall\left( \bigwedge_{\forall j \,|\, {m^-(x,\, i) < j < m^+(x,\, i) \atop x_j = r_j}} y_j \right)\Bigg|\, y_j = r_j, \neg r_j$$

where

$$\bigwedge_{\forall j \,|\, {m^-(x,\, i) < j < m^+(x,\, i) \atop x_j = r_j}} y_j$$

is a general description of a combination of statements. As an example, consider the path $< r_1, r_2, \neg r_3, r_4, r_5 >$. We have $m^-(x, i) = 1$, $m^+(x, i) = 5$ and $x_j = r_j$, $j \neq 3$. The general description of a combination is $(y_2 \wedge y_4)$ where

$$\forall (y_2 \wedge y_4) \,|\, y_j = r_j, \neg r_j$$

yields the set of all combinations, $\{r_2 \wedge r_4, r_2 \wedge \neg r_4, \neg r_2 \wedge r_4, \neg r_2 \wedge \neg r_4\}$. The contribution from each combination depends on the positive evidence for that combination, the term

$$\prod_{\forall k \,|\, {m^-(x,\, i) < k < m^+(x,\, i) \atop x_k = r_k}} \begin{cases} p_k, & y_k = r_k \\ 1 - p_k, & y_k = \neg r_k \end{cases},$$

and the negative evidence given by the following necessary conditions for that combination.

The first condition is that all internal vertices must be visited. Hence, for each sequence of vertices among the internal vertices, that are not stated to be true in this combination, Figure 6, we must block all forbidden edges. These are edges from a vertex $v_i$ to a vertex $v_j$ where $v_i$ is in the sequence or the last vertex before the sequence, $v_j$ is in the sequence or the first vertex after the sequence and where there is a vertex $v_k$ such that $v_k$ is in the sequence, i.e., the sequence of internal vertices not stated to be true in the present combination, and $i < k < j$. It is accomplished by the term

$$\left( \prod_{\forall k \left| \substack{m^-(x, i) < k < m^+(x, i) \\ x_k = r_k \\ y_k = \neg r_k} \right.} \left( \prod_{\forall m \left| \substack{k < m \leq \min(m^-(y, i, k), m^+(x, i)) \\ x_m = r_m} \right.} q_{m^+(x, i, k), m} \right) \right)$$

where $m^+(x, i, k)$ is the last vertex in the path before the vertex $v_k$ not stated to be true and $\min(m^-(y, i, k), m^+(x, i))$ is the first vertex after the sequence of vertices not stated to be true. As an example of the first condition, consider again the path $< r_1, r_2, \neg r_3, r_4, r_5 >$ now for the combination $(y_2 \wedge y_4) = (\neg r_2 \wedge \neg r_4)$. In Figure 7 the necessary edges are blocked. These are the edges $v_1$ to $v_4$, $v_1$ to $v_5$, and $v_2$ to $v_5$. Vertices $v_1$ and $v_2$ are in or the last vertex before the sequence, $v_4$ and $v_5$ are in or the first vertex after the sequence, and there is at least one vertex between the two vertices of the edge, in these cases $v_2$, $v_2$ and $v_4$, and vertex $v_4$ respectively.

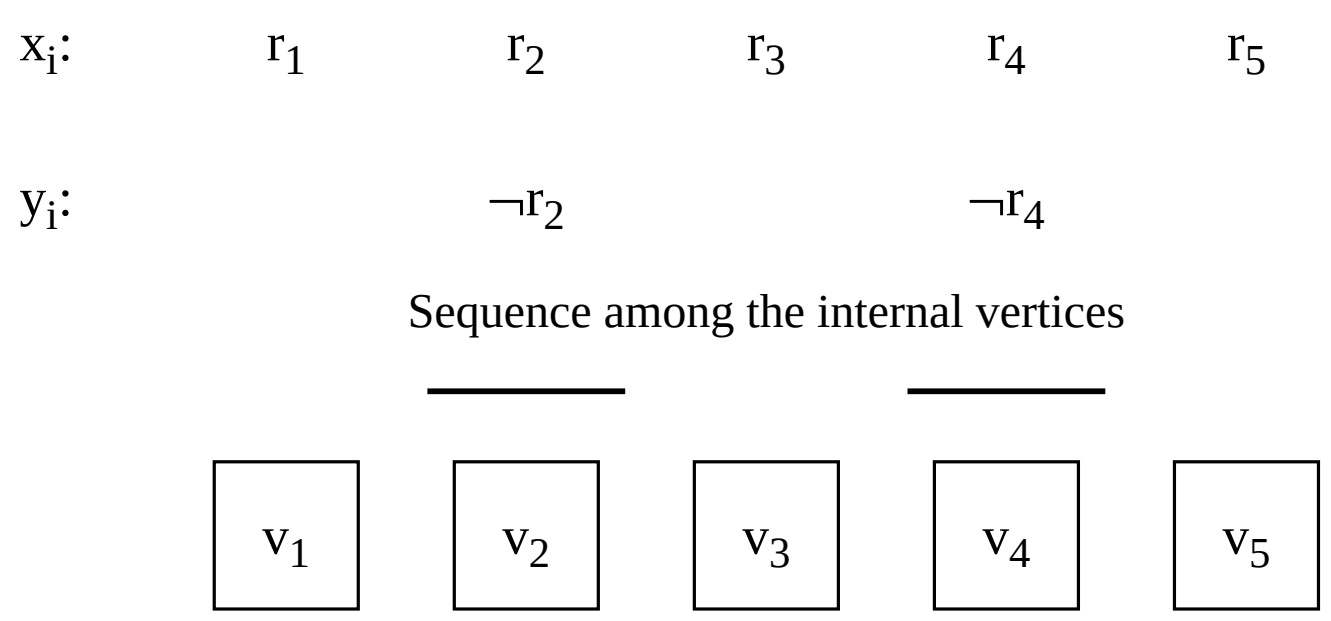

**Figure 6.** Vertices $v_2$ and $v_4$ form a sequence because $v_3$ is not in the path.

The second and final condition will assure that, between the first and last vertex of the path, no vertices other than those in the path are visited. The evidence against every edge from a vertex $v_i$ to a vertex $v_j$ where $v_i$ is included in the path, $v_j$, $i < j < s_n$, is not included in the path and where there are no internal vertices $v_k$, $i < k < j$, that are stated to be true to the path in the present combination, Figure 8, is to some degree committed in favor of the path and is for the remainder uncommitted. However, for each combination of statements for the internal vertices we will have support for the path from all combinations of statements for sets of edges from earlier vertices to an internal vertex, Figure 9, given correct conditions for the edges from this internal vertex. The evidence against set of edges, from vertices $v_i$ to a vertex $v_j$ where there for each $v_i$ are no internal vertices $v_k$, $i < k < j$, stated to be true, is considered to be true if all edges in the set are blocked. Hence, for each combination of statements for internal vertices we will sum up the contribution from all combinations of statements for the set of edges to these vertices. If the evidence against the set of edges to one of these vertices is not stated to be true, then we should sum up the contribution from all those alternatives of the edges to that vertex where at least the evidence of one of the edges is not stated to be true. An example of when the evidence against the set of edges to one vertex is not stated to be true, i.e., when all edges in the set are not blocked, is the three alternatives of the second combination of set of edges to vertex 3 in Figure 9. We must also take into account the necessary conditions on the edges from that vertex. However, because the conditions are the same for all alternatives when at least one of the edges to the vertex is not blocked, as with the three alternatives for the second combination in Figure 9, we are able to view all these alternatives in a set of edges not stated to be true as one generalized edge to the vertex that is not stated to be true for the present combination of internal vertices. The necessary conditions are that the edges from the vertex to all internal vertices not stated to be true in a subsequent sequence and to the first vertex after the sequence are blocked (Figure 10). Its contribution is

$$\left(1 - \prod_{\forall u \,\middle|\, \substack{\max(m^{+}(y,\, i,\, t),\, m^{-}(x,\, i)) \leq u < t \\ (z_u = \neg r_u) \vee (x_u = r_u)}} q_{u,\, t}\right),$$

where $t$ is the index of the vertex not included in the path, $u < t < n_s$, $z_u$ marks whether or not all edges from vertices $v_j$ in the path to vertex $v_u$, $j < u < s_n$, where there are no internal vertices $v_k$, $j < k < u$, stated to be true, are blocked. The necessary condition on the edges from vertex $v_t$ are:

$$\prod_{\forall v \,\middle|\, \substack{t < v \leq \min(m^{-}(y,\, i,\, t),\, m^{+}(x,\, i)) \\ x_v = r_v}} q_{t,\, v}.$$

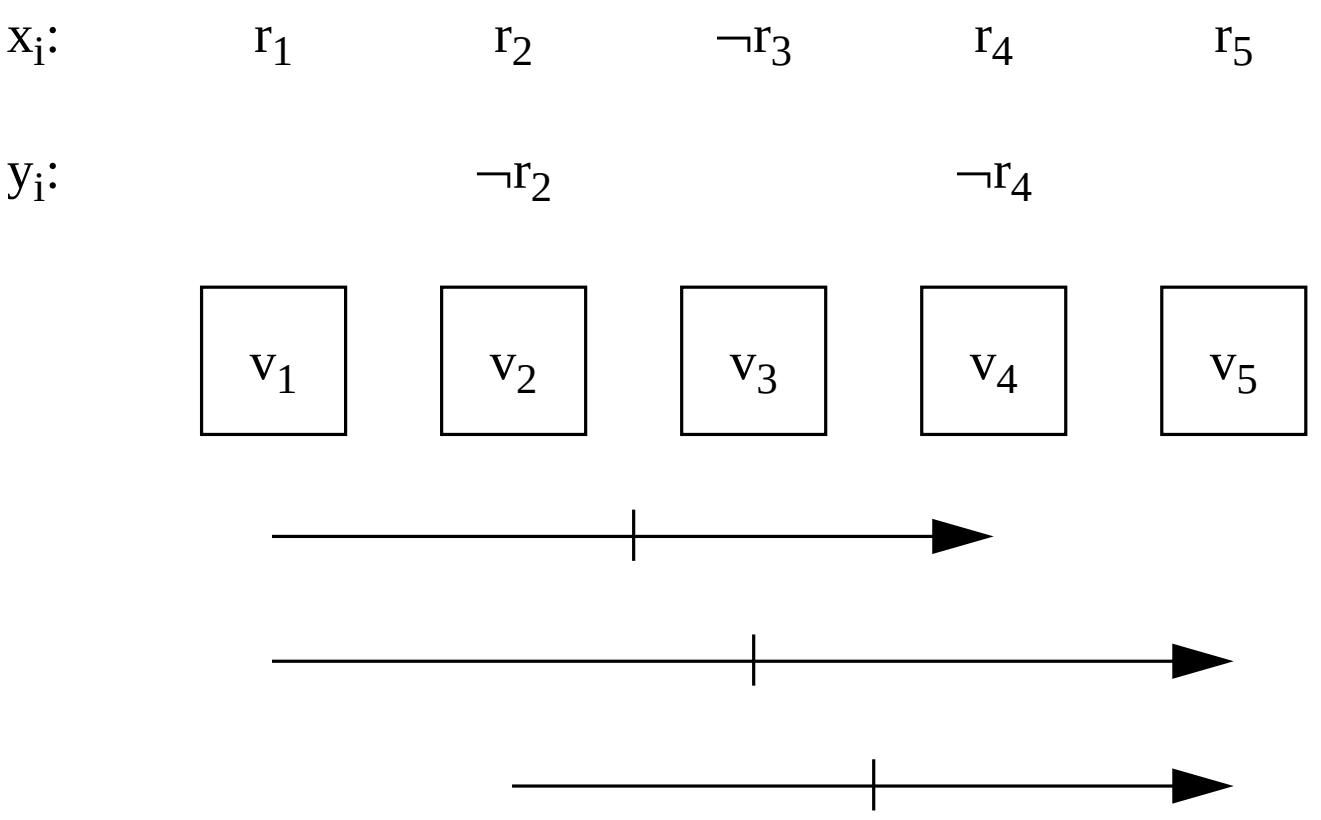

**Figure 7.** Example of the first condition for the combination $(\neg r_2 \wedge \neg r_4)$.

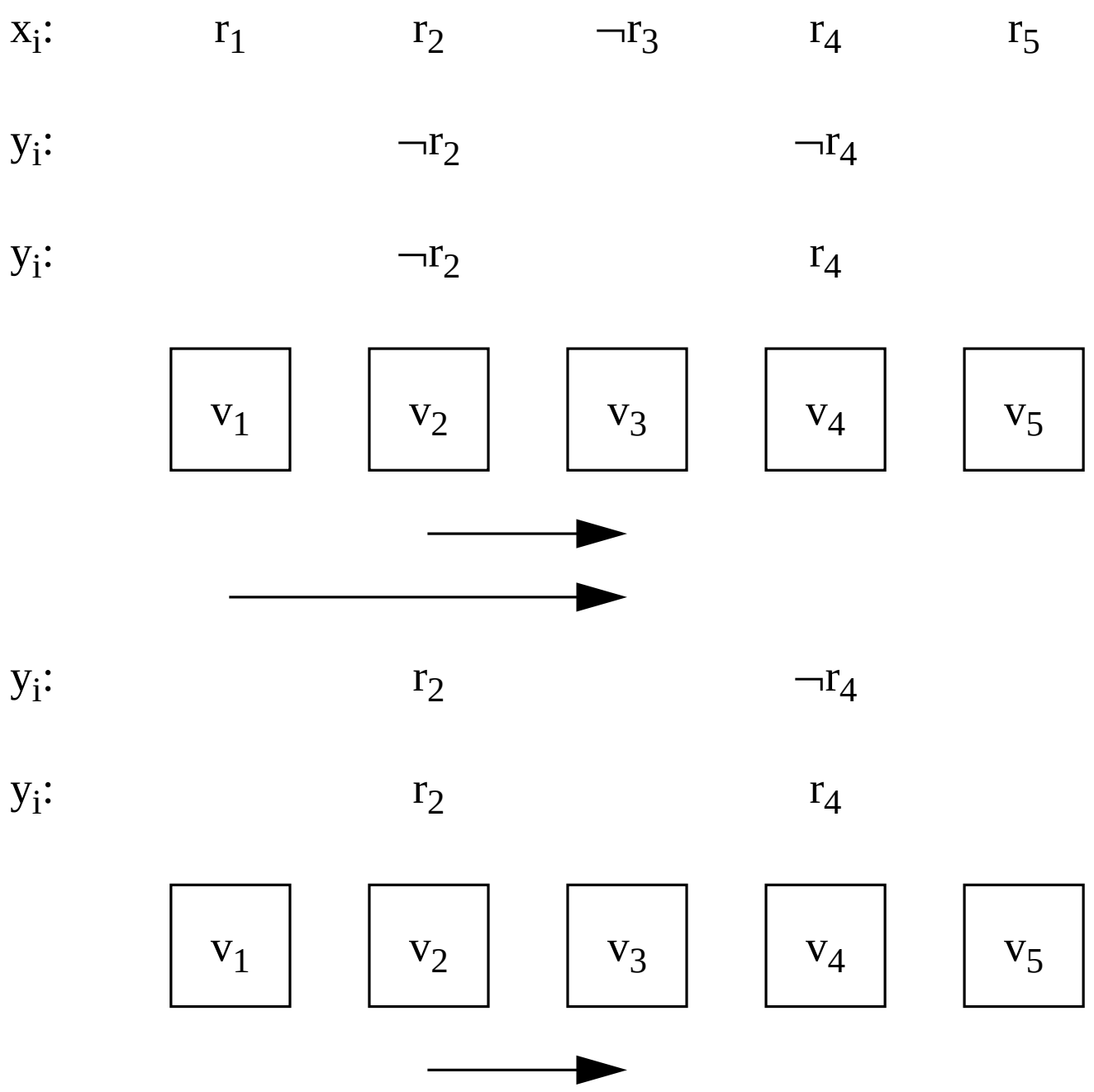

**Figure 8.** Possible edges to $v_3$ for different combinations.

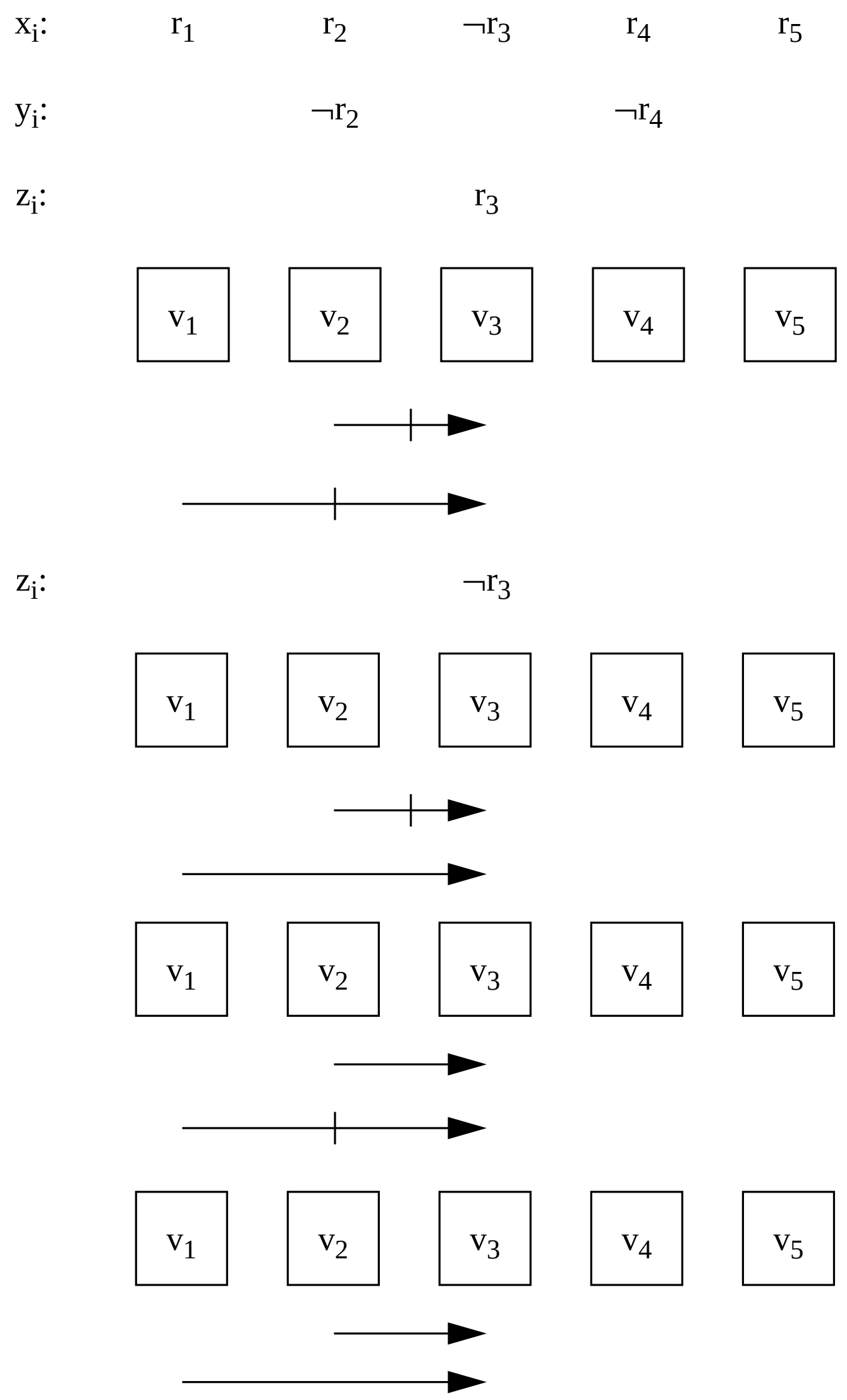

**Figure 9.** Two different states, $z_3$ (= $r_3$, $\neg r_3$), with one and three alternatives.

The final alternative that all edges to the vertex are blocked, as in the first combination of set of edges in Figure 9, involve no conditions. Its

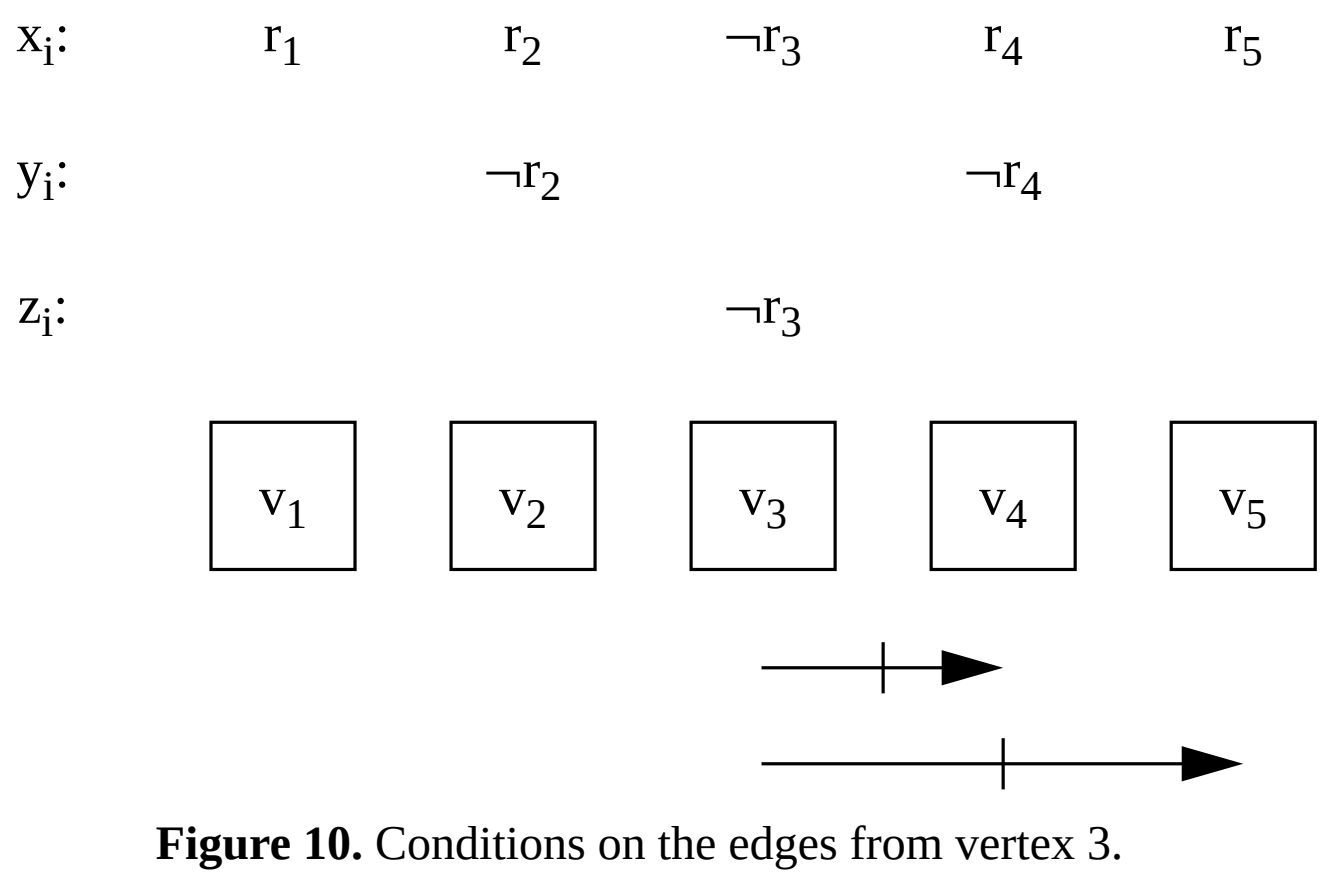

**Figure 10.** Conditions on the edges from vertex 3.

contribution is:

$$\prod_{\substack{\forall u \mid \max(m^{+}(y,\,i,\,t),\, m^{-}(x,\,i)) \le u < t \\ (z_u = \neg r_u) \vee (x_u = r_u)}} q_{u,\,t}.$$

As an example of the second condition, consider the path < $r_1$, $\neg r_2$, $\neg r_3$, $r_4$, $r_5$> for the combination of statements for the internal vertex $y_4 = \neg r_4$ and the combination of statements for set of edges ( $z_2 \wedge z_3$) = ( $\neg r_2 \wedge \neg r_3$), Figure 11. That is, nothing speaks in favor of vertex 4. Furthermore, consider the edges where there are no internal vertices, between the vertices of the edge, stated to be true in the present combination. There is nothing that speaks against that there is at least one of these edges from a vertex in the path to vertices 2 and 3 respectively that is not blocked. If there is an edge to vertex 2 then it must be coming from vertex 1. The necessary condition is that all edges from vertex 2 to all internal vertices not stated to be true in a subsequent sequence and the first vertex after the sequence are blocked. Because vertex 4 is not stated to be true in this combination it is necessary to block the edges $v_2$ to $v_4$ and $v_2$ to $v_5$. There are two different edges to vertex 3, $v_1$ to $v_3$ and $v_2$ to $v_3$. At least one of these two should not be blocked. This gives us three different alternatives, $v_1$ to $v_3$ and not $v_2$ to $v_3$, not $v_1$ to $v_3$ and $v_2$ to $v_3$ and finally $v_1$ to $v_3$ and $v_2$ to $v_3$. The corresponding term becomes $(1 - q_{13}) \cdot q_{23} + q_{13} \cdot (1 - q_{23}) + (1 - q_{13}) \cdot (1 - q_{23})$, rewritten as $(1 - q_{13}) + q_{13} \cdot (1 - q_{23})$ it is understood as $v_1$ to $v_3$ or if not $v_1$ to $v_3$ then $v_2$ to $v_3$, as described in Figure 11. Rewriting the term as $1 - q_{13} \cdot q_{23}$ can be interpreted as one generalized edge to vertex 3 whose evidence is not stated to be true. This is the way it is rewritten in the algorithm. The condition for vertex 3 is of the same type as for vertex 2, here that the edges $v_3$ to $v_4$ and $v_3$ to $v_5$ are blocked.

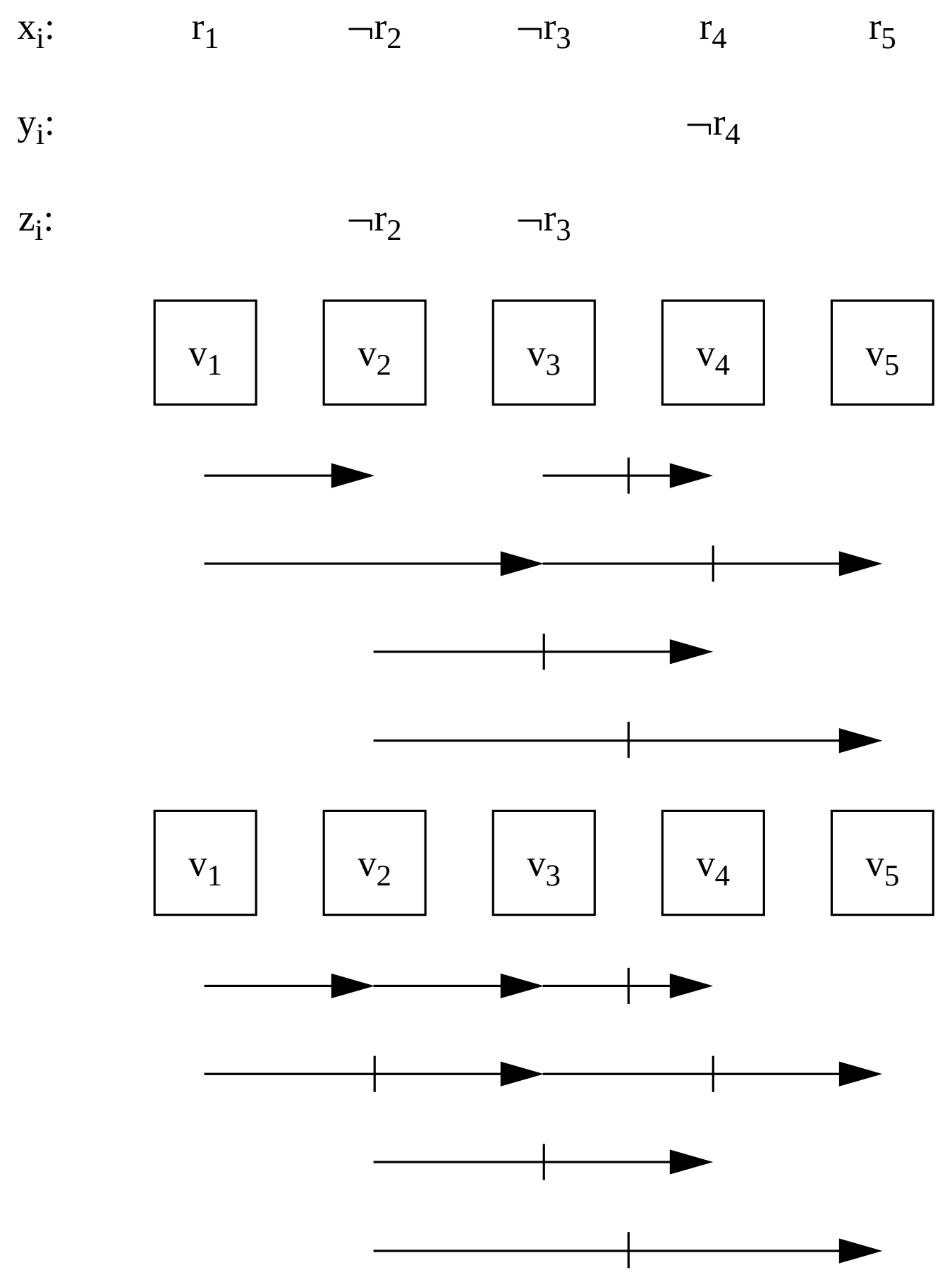

**Figure 11.** The second condition for the combination $y_4 = \neg r_4$, $(z_2 \wedge z_3) = (\neg r_2 \wedge \neg r_3)$.

6.2.3. THE ALGORITHM FOR SUPPORT The algorithm for support can then be summarized as

$$\forall x_i \big| x_i \in \{r_i, \neg r_i\} . \mathrm{Spt}(<x_1, x_2, \ldots, x_n>)$$

$$= \frac{1}{1-k_n} \cdot p_{m^-(x,i)} \cdot \begin{cases} p_{m^+(x,i)}, m^-(x,i) \neq m^+(x,i) \\ 1, m^-(x,i) = m^+(x,i) \end{cases}$$

$$\cdot \left( \prod_{\forall i | (x_i = \neg r_i)} (1-p_i) \right) \cdot \left( \prod_{\forall j | 1 \le j < m^-(x,i)} q_{j, m^-(x,i)} \right)$$

$$\cdot \left( \prod_{\forall j | m^+(x,i) < j \le n} q_{m^+(x,i), j} \right) \cdot \left( \prod_{\forall j \big|^{j \neq m^+(x,i)}_{x_j = r_j}} 1 - q_{j, m^-(x,i,j)} \right)$$

$$\cdot \left( \sum_{\forall \left( \bigwedge_{\forall j \big|^{m^-(x,i) < j < m^+(x,i)}_{x_j = r_j}} y_j \right) \Big| y_j = r_j, \neg r_j} \right.$$

$$\left. \left( \prod_{\forall k \big|^{m^-(x,i) < k < m^+(x,i)}_{x_k = r_k}} \begin{cases} p_k, y_k = r_k \\ 1 - p_k, y_k = \neg r_k \end{cases} \right) \right)$$

$$\cdot \left( \prod_{\forall k \,\middle|\, \substack{m^-(x,\, i) \,<\, k \,<\, m^+(x,\, i) \\ x_k \,=\, r_k \\ y_k \,=\, \neg r_k}} \left( \prod_{\forall m \,\middle|\, \substack{k \,<\, m \,\le\, \min(m^-(y,\, i,\, k),\, m^+(x,\, i)) \\ x_m \,=\, r_m}} q_{m^+(x,\, i,\, k),\, m} \right) \right)$$

$$\cdot \left( \sum_{\forall \left( \bigwedge_{\forall s \,\middle|\, \substack{m^-(x,\, i) \,<\, s \,<\, m^+(x,\, i) \\ x_s \,=\, \neg r_s}} z_s \right) \middle|\, z_s \,=\, r_s,\, \neg r_s} \left( \prod_{\forall t \,\middle|\, \substack{m^-(x,\, i) \,<\, t \,<\, m^+(x,\, i) \\ x_t \,=\, \neg r_t}} \begin{cases} \xi, & z_t = r_t \\ \psi, & z_t = \neg r_t \end{cases} \right) \right)$$

where

$$\xi = \prod_{\forall u \,\middle|\, \substack{\max(m^+(y,\, i,\, t),\, m^-(x,\, i)) \,\le\, u \,<\, t \\ (z_u \,=\, \neg r_u) \,\vee\, (x_u \,=\, r_u)}} q_{u,\, t}$$

and

$$\psi = \left( 1 - \prod_{\forall u \,\middle|\, \substack{\max(m^+(y,\, i,\, t),\, m^-(x,\, i)) \,\le\, u \,<\, t \\ (z_u \,=\, \neg r_u) \,\vee\, (x_u \,=\, r_u)}} q_{u,\, t} \right) \cdot \prod_{\forall v \,\middle|\, \substack{t \,<\, v \,\le\, \min(m^-(y,\, i,\, t),\, m^+(x,\, i)) \\ x_v \,=\, r_v}} q_{t,\, v}$$

and $k_n$ is the conflict in the $n$-vertex graph.

### 6.3. Conflict

The conflict indicates the amount of the total mass consisting of contradictory evidences, i.e., evidences whose intersection is the empty set, ∅. This means that we actually compute the support for ∅, but as we do not want to assign any belief to an impossible event, this is denoted conflict,

$$\text{Conf} = \sum_{A = \varnothing} m(A).$$

6.3.1. EXPLAINING THE ALGORITHM FOR CONFLICT In our case the conflict arises when combining the evidences $e_{ij}$ with the evidences concerning vertices $v_i$, ..., $v_j$. The calculations are based on the formula:

$$\text{Conf}(\tilde{e}_1, \tilde{e}_2, \ldots, \tilde{e}_{n+1}) = \text{Conf}(\tilde{e}_1, \tilde{e}_2, \ldots, \tilde{e}_n) + \text{Conf}(\tilde{e}_1 \oplus \tilde{e}_2 \oplus \ldots \oplus \tilde{e}_n, \tilde{e}_{n+1}),$$

where $\tilde{e}_i$ are arbitrary evidences and $\tilde{e}_i \oplus \tilde{e}_j$ is the combined evidence from $\tilde{e}_i$ and $\tilde{e}_j$. We here in fact mean the basic probability assignment for the evidences, but for simplicity we use the denotation for evidence. The formula above means that when we add new evidences to already combined evidences the new conflict is obtained as the sum of the earlier conflict and a contribution from the new evidences. The conflict can never decrease when bringing in new evidences. For the sake of clarity we assume that the combination of evidences take position stepwise in the following order:

$$e_1 \oplus e_2 \oplus e_{12} \oplus e_3 \oplus e_{23} \oplus e_{13} \oplus e_4 \oplus \ldots \oplus e_n \oplus e_{n-1n} \oplus \ldots \oplus e_{1n}.$$

The positive evidences $e_i$ are brought into the combination in increasing order of $i$, but between the $e_i$ all negative evidences $e_{ij}$ are regarded in such a way that $e_k$ is followed by all $e_{ik}$ where $i < k$. This means that the $e_i$ never give rise to any conflict when they are brought into the combination which on the other hand the $e_{ij}$ do. We denote the contribution from $e_{ij}$ to the already existing conflict by $k_{ij}$, i.e.,

$$k = \sum_{j=2}^{n} \sum_{i=1}^{j-1} k_{ij}.$$

Let us look at what happens when we bring in the specific evidence $e_{ij}$ to the combination. As mentioned earlier this may give a conflict based on earlier evidences.

Let $S_{ij} = < x_i, x_{i+1}, \ldots, x_j >$ where $x_i = r_i$, $x_j = r_j$ and $x_k = \neg r_k$, $i < k < j$. $e_{ij}$ speaks against the subpath $S_{ij}$ to the degree $q_{ij}$. The earlier evidences speaks for this subpath to the degree

$$\frac{\text{Sup}^*(S_{ij})}{(1 - q_{ij})},$$

where Sup*($S_{ij}$) is calculated as described earlier, but the computation of Sup*( $S_{ij}$) was then based on the evidence $e_{ij}$ itself, which is not relevant here. The influence of $e_{ij}$ on Sup*($S_{ij}$) is neglected by division with its contribution (1 - $q_{ij}$). The total conflict caused by $e_{ij}$ is consequently $c_{ij} = q_{ij} \cdot \mathrm{Sup}(S_{ij}) / (1 - q_{ij})$ but this conflict is not equal to the contribution $k_{ij}$ because a part of $c_{ij}$ is already taken into account by the calculated conflict based on the earlier evidences. This means that $c_{ij}$ has to be reduced in the following way. The total conflict before $e_{ij}$ is

$$\sum_{h=2}^{j-1}\sum_{k=1}^{h-1} k_{kh} + \sum_{k=i+1}^{j-1} k_{kj}.$$

This expression can be written as a sum of the four terms:

$$\sum_{h=2}^{i-1}\sum_{k=1}^{h-1} k_{kh} + \sum_{k=1}^{i-1} k_{ki} + \sum_{h=i+1}^{j-1}\sum_{k=1}^{h-1} k_{kh} + \sum_{k=i+1}^{j-1} k_{kj}.$$

Let us consider the first term:

$$\sum_{h=2}^{i-1}\sum_{k=1}^{h-1} k_{kh}.$$

This conflict is only based on the evidences concerning vertices before $v_i$, therefore we may have a conflict based on these evidences at the same time as we have a conflict only based on evidences from vertex $v_i$ and forward. The new contribution to the conflict, $k_{ij}$, must not contain the earlier conflict. Hence, $c_{ij}$ is reduced by the term

$$c_{ij} \cdot \sum_{h=2}^{i-1}\sum_{k=1}^{h-1} k_{kh},$$

which is the degree to which we have conflict in both.

For the second term,

$$\sum_{k=1}^{i-1} k_{ki},$$

the reasoning is almost the same as for the first term with the difference that in the expression for the simultaneous conflict,

$$c_{ij} \cdot \sum_{k=1}^{i-1} k_{ki},$$

the support $p_i$ for the evidence $e_i$ occurs in both the factor $c_{ij}$ and the factors $k_{ij}$, which must not be the case when they are regarded simultaneously, therefore the expression has to be divided by $p_i$, i.e., the

reducing term based on:

$$\sum_{k=i+1}^{j-1} k_{ki},$$

equals:

$$\frac{c_{ij}}{p_i} \cdot \sum_{k=i+1}^{j-1} k_{ki}.$$

For the last two terms in the sum above, every $k_{kh}$ is based on at least one evidence $e_k$ concerning a vertex between $v_i$ and $v_j$. This means that it is impossible to have a conflict based on the evidence $e_{ij}$ at the same time as we state a vertex between $v_i$ and $v_j$ to be true, so the last two terms in the sum do not contain any part of the conflict $c_{ij}$ and do not contribute to the reduction.

This means that

$$\begin{aligned} k_{ij} &= c_{ij} \cdot \left(1 - \sum_{h=2}^{i-1} \sum_{k=1}^{h-1} k_{kh} - \frac{1}{p_i} \cdot \sum_{k=1}^{i-1} k_{ki}\right) \\ &= \frac{q_{ij}}{(1-q_{ij})} \cdot \left(1 - \sum_{h=2}^{i-1} \sum_{k=1}^{h-1} k_{kh} - \frac{1}{p_i} \cdot \sum_{k=1}^{i-1} k_{ki}\right). \end{aligned}$$

This is true for $i \geq 3$.

If $i = 1$ the $c_{ij}$ do not have to be reduced because in this case the reasoning is the same as for the last two terms.

For $i = 2$ the reducing factor is

$$\frac{1}{p_i} \cdot \sum_{k=1}^{i-1} k_{ik},$$

and

$$k_{ij} = \frac{1}{1-q_{ij}} \cdot \left(1 - \frac{1}{p_i} \cdot \sum_{k=1}^{i-1} k_{ki}\right).$$

6.3.2. THE ALGORITHM FOR CONFLICT The conflict, $k_n$, of a graph with $n$ vertices can be calculated as

$$k_n = \begin{cases} \sum_{i=1}^{n-1} \sum_{j=i+1}^{n} k_{ij}, & n > 1 \\ 0, & n = 1 \end{cases}$$

where

$$k_{ij} = \begin{cases} \left(k_{1j-i+1}\sigma^{\mathrm{I}}_{i-1}\right)\sigma^{\mathrm{II}}_{i}, & i > 1 \\ \dfrac{q_{1j}}{1-q_{1j}} \cdot \mathrm{Spt}^{*}(<r_1, \neg r_2, \neg r_3, \ldots, \neg r_{j-1}, r_j>), & i = 1 \end{cases}$$

and $\sigma^{\mathrm{I}}_{i}$ and $\sigma^{\mathrm{II}}_{i}$ are the substitutions

$$\sigma^{\mathrm{I}}_{i} = \forall m, n. \{p_m / p_{m+i}, q_{mn} / q_{m+in+i}\},$$

$$\sigma^{\mathrm{II}}_{i} = \begin{cases} \left\{ p_i \Big/ \left( p_i - p_i \cdot \sum_{m=1}^{i-2} \sum_{n=m+1}^{i-1} k_{mn} - \sum_{m=1}^{i-1} k_{mi} \right) \right\}, & i > 2 \\ \{p_i / (p_i - k_{12})\}, & i = 2 \end{cases}$$

and Spt*( < $r_1$, $\neg r_2$, $\neg r_3$, ..., $\neg r_{j-1}$, $r_j$ > ) is the unnormalized support.

## 7. COMPLEXITY

The time complexity of the classic algorithm is of course such that using it in any real-time application is out of the question. But even when one is using it for symbolic precalculations one runs into problems, as seen in Figure 12. The space complexity of the classic algorithm, Figure 12, should, however, not be interpreted as the size of the data to be handled by an application, but rather the size of the expressions that ought to be simplified by some algebraic system.

Neither can the new algorithm be used in real-time applications for anything but the smallest problems, but it is feasible to use it for other applications as well as for symbolic precalculations. On today's supercomputers the new algorithm can manage graphs of up to 36 vertices in size, i.e., up to 666 evidences with $|\Theta| = 2^{36}$, when calculating support for one single instance of the frame (1 Gflops for 10 minutes) as compared to only six vertices for the classic algorithm.

| | The classic algorithm | The new algorithm |
|---|---|---|
| Time complexity | $O(\lvert\Theta\rvert^{log\lvert\Theta\rvert})$ | $O(\lvert\Theta\rvert \cdot log\lvert\Theta\rvert)$ |
| Space complexity | $O(\lvert\Theta\rvert^{log\lvert\Theta\rvert} \cdot log^2\lvert\Theta\rvert)$ | $O(\lvert\Theta\rvert \cdot log\lvert\Theta\rvert)$ |

**Figure 12.** Complexity of the classic and new algorithms.

### 7.1. The Classic Algorithm

Assuming that there are $n$ vertices in the graph, the time complexity of $\mathrm{Bel}_p$ is $O(2^n)$ and the space complexity is $O(n \cdot 2^n)$. When there are $n$ vertices there are $\frac{1}{2} \cdot n \cdot (n-1)$ edges. Thus, the time complexity of $\mathrm{Bel}_n$ becomes $O(2^{(n^2)})$ with a space complexity of $O(n^2 \cdot 2^{(n^2)})$. The time complexity of $\mathrm{Bel}_p \oplus \mathrm{Bel}_n$ will then be $O(2^{(n^2)})$ and the space complexity $O(n^2 \cdot 2^{(n^2)})$, or when measured in the size of the frame $O(|\Theta|^{\log|\Theta|})$ and $O(|\Theta|^{\log|\Theta|} \cdot \log^2|\Theta|)$ respectively.

### 7.2. The New Algorithm

The unnormalized plausibility for a single path can be calculated in linear time. The time complexity of the unnormalized support for a single path is far worse, being determined by the summation over the three last factors which are $O(n \cdot (\frac{3}{2})^n)$, $O(n \cdot (\frac{3}{2})^n)$ and $O(n \cdot 2^n)$ respectively. Thus, the time complexity of calculating the unnormalized support for a single instance of the frame becomes $O(n \cdot 2^n)$. If we assume that the unnormalized support for one particular path for each graph size is already calculated, then the time complexity of calculating the conflict will be $O(2^n)$, otherwise we must calculate the unnormalized support for these paths yielding a time complexity for the conflict of $O(n \cdot 2^n)$. Thus, the time complexity of calculating support and plausibility for each path is $O(n \cdot 2^n)$, or when measured in the size of the frame $O(|\Theta| \cdot \log|\Theta|)$. Presumably we can use domain knowledge to substantially restrict the number of credible scenarios.

The space complexity, when calculating support and plausibility symbolically, is equal to the time complexity.

## 8. CONCLUSIONS

We have presented an algorithm that makes it feasible to precalculate support and plausibility symbolically for completely specified paths through a complete directed acyclic graph. One problem when reasoning about completely specified paths, i.e., paths where $\forall i . x_i \neq \theta_i$, is that for larger graphs there might be a large number of quite similar paths with equally low support and plausibility. The average characterization of these paths may then be lost. If there is no completely specified path which stands out from the analysis, this would make the calculation useless for decision support. We might therefore also be interested in reasoning about incompletely specified paths, i.e., subparts of paths.

## ACKNOWLEDGMENT

We would like to thank Stefan Arnborg and Per Svensson for their helpful comments regarding this paper.